\documentclass{egpubl}
\usepackage{eg2026}
 
\ConferencePaper        
\usepackage[T1]{fontenc}
\usepackage{dfadobe}  

\usepackage{cite}  
\BibtexOrBiblatex
\electronicVersion
\PrintedOrElectronic
\ifpdf \usepackage[pdftex]{graphicx} \pdfcompresslevel=9
\else \usepackage[dvips]{graphicx} \fi

\usepackage{egweblnk} 

\makeatletter
\providecommand{\p@copyrightTextLong}{}
\providecommand{\p@copyrightTextShort}{}
\providecommand{\p@copyrightTextShortEven}{}
\providecommand{\p@copyrightTextTitPag}{}
\makeatother
\usepackage{lineno}
\usepackage{mathrsfs}
\usepackage{tikz}
\usetikzlibrary{arrows.meta}
\usepackage{overpic}
\usepackage{pict2e}
\usepackage{dsfont}
\usepackage[most]{tcolorbox}
\usepackage[outline]{contour}
\usepackage{standalone}
\usepackage[normalem]{ulem}
\usepackage{xcolor}

\usepackage{hyperref}       
\usepackage{microtype}      
\usepackage[capitalize]{cleveref}
\usepackage{xspace}         
\usepackage{mathtools}      
\usepackage{nicefrac,xfrac} 
\usepackage{mathtools}

\usepackage{stmaryrd}  
\usepackage{newfloat}  
\usepackage{savesym}
\usepackage{algorithm}
\usepackage{algorithmicx}
\usepackage[noend]{algpseudocode} 
\algrenewcommand\textproc{}
\usepackage{graphicx}
\usepackage{grffile}  
\usepackage{soul}
\usepackage{wrapfig}
\usepackage{bbm}
\usepackage{flushend}
\usepackage{tabularx}
\usepackage{ctable}            
\usepackage[a]{esvect}           
\usepackage{multirow}
\usepackage{pdfpages}

\usepackage{wrapfig}
\usepackage{textcomp}    
\usepackage{enumitem}    

\usepackage{caption}

\usepackage{amssymb, amsmath, commath}

\newcommand{\Ignore}[1]{} 

\newcommand{\Sample}{\mathbf{x}}
\newcommand{\Samplet}{\mathbf{x}_t}

\newcommand{\SampleT}{\mathbf{x}_T}
\newcommand{\Samples}{\mathbf{x}_s}
\newcommand{\Samplezero}{\mathbf{x}_0}
\newcommand{\SamplezeroApprox}{\hat{\mathbf{x}}_0}

\newcommand{\Loss}{\mathcal{L}}
\newcommand{\OurLossFn}{||\Model - \SigmatTensor \Noise||^2}
\newcommand{\Model}{f_\theta(\boldsymbol{x}_t, t)}

\newcommand{\SignalCoeff}{\gamma}

\newcommand{\SigmatScalar}{\sigma_t}
\newcommand{\SigmasScalar}{\sigma_s}
\newcommand{\SigmatTensor}{\boldsymbol{\sigma}_t}

\newcommand{\Noise}{\boldsymbol{\epsilon}_t}
\newcommand{\NoiseApprox}{\hat{\boldsymbol{\epsilon}}_t}

\newcommand{\Identity}{\boldsymbol{I}}
\newcommand{\NormalDist}{\mathcal{N}}

\newcommand{\AlphatBar}{\bar{\alpha_t}}
\newcommand{\AlphasBar}{\bar{\alpha_s}}
\newcommand{\AlphaCumProd}{\prod_{i=1}^{t} \alpha_{i}}
\newcommand{\SqrtAlphatBar}{\sqrt{\AlphatBar}}
\newcommand{\SqrtAlphasBar}{\sqrt{\AlphasBar}}
\newcommand{\SqrtOneMinAlphatBar}{\sqrt{1 - \AlphatBar}}

\newcommand{\Alphat}{\SignalCoeff_t}
\newcommand{\Alphas}{\SignalCoeff_s}
\newcommand{\Alphatovers}{\SignalCoeff_{t|s}}

\newcommand{\Sigmatovers}{\sigma_{t|s}}

\newcommand{\Meanttos}{\boldsymbol{\mu}_{t \rightarrow s}}
\newcommand{\Sigmattos}{\sigma_{t \rightarrow s}}
\newcommand{\SigmattosTensor}{\boldsymbol{\sigma}_{t \rightarrow s}}

\newcommand{\DiffusionCoeffTensor}{\mathbf{c}(\mathbf{x},t)}
\newcommand{\DiffusionCoeffTensorNonTimeDependent}{\mathbf{c}_0(\mathbf{x})}
\newcommand{\DiffusionCoeffTensorZero}{\mathbf{c}(\mathbf{x},0)}

\newcommand{\DiffusionCoeffTensorXs}{\mathbf{c}(\mathbf{x}_s,s)}
\newcommand{\DiffusionCoeffTensorXZero}{\mathbf{c}(\mathbf{x}_0,t)}

\newcommand{\LaplaceSample}{\Delta \mathbf{x}}

\newcommand{\PeronaMalikFilterReciprocal}{\sqrt{1 + \frac{\ImageGradient}{\lambda}}}

\newcommand{\PeronaMalikFilterTimeDependentReciprocal}{\sqrt{1 + \frac{\ImageGradient}{\EdgeSensitivityTimeVarying}}}

\newcommand{\ImageGradient}{||\nabla \Sample||}

\newcommand{\StartImageGradient}{||\nabla \Samplezero||}

\newcommand{\EdgeSensitivity}{\lambda}
\newcommand{\EdgeSensitivityMin}{\lambda_{min}}
\newcommand{\EdgeSensitivityMax}{\lambda_{max}}
\newcommand{\EdgeSensitivityTimeVarying}{\lambda(t)}
\newcommand{\EdgeSensitivityTimeVaryingTimeDerivative}{\lambda\sp{\prime}(t)}
\newcommand{\TransitionFn}{\tau(t)}
\newcommand{\TransitionFnTimeDerivative}{\tau\sp{\prime}(t)}

\newcommand{\TransitionPt}{t_{\Phi}}

\newcommand{\NoiseCoeffNumerator}{b}

\newcommand{\SigmaOursGeneral}{\frac{\NoiseCoeffNumerator} \DiffusionCoeffTensorXZero}

\newcommand{\SigmaOursTimeVaryingGeneral}{\frac{\NoiseCoeffNumerator}{(1 - \TransitionFn) \DiffusionCoeffTensorXZero + \TransitionFn}}

\newcommand{\OurAlphaToverSSqrd}{\frac{\AlphatBar}{\AlphasBar}}
\newcommand{\OurAlphaToverS}{\frac{\SqrtAlphatBar}{\SqrtAlphasBar}}

\newcommand{\OurBackwardPosteriorVariance}{\left(\OurBackwardPosteriorVarianceTermOne + \OurBackwardPosteriorVarianceTermTwo \right)^{-1}}

\newcommand{\AuxiliaryOne}{\boldsymbol{\sigma}^2(t)}
\newcommand{\AuxiliaryOneS}{\boldsymbol{\sigma}^2(s)}

\newcommand{\OurBackwardPosteriorVarianceTermOne}{\frac{1}{\AuxiliaryOne}}
\newcommand{\OurBackwardPosteriorVarianceTermTwo}{\frac{\OurAlphaToverSSqrd}{\AuxiliaryOne - \OurAlphaToverSSqrd \AuxiliaryOneS}}

\newcommand{\OurBackwardPosteriorMeanTermOne}{\frac{\OurAlphaToverS}{\AuxiliaryOne - \OurAlphaToverSSqrd \AuxiliaryOneS} \Samplet}
\newcommand{\OurBackwardPosteriorMeanTermTwo}{\frac{\SqrtAlphasBar}{\AuxiliaryOneS} \Samplezero}
\newcommand{\OurBackwardPosteriorMean}{\Sigmattos^2 \left(\OurBackwardPosteriorMeanTermOne + \OurBackwardPosteriorMeanTermTwo \right)}

\definecolor{myyellowish}{rgb}{1, 0.647, 0}
\definecolor{myredish}{rgb}{1, 0, 0}
\definecolor{mybluish}{rgb}{0.1216, 0.4667, 0.7059}

\newcommand\timepoint t
\newcommand\prevtimepoint r

\newcommand\diffusiondrift\beta
\newcommand\diffusioncoefficient\varsigma

\newcommand\point x
\newcommand\datadistribution\mu
\newcommand\weighting\alpha
\newcommand\finaltime T
\newcommand\score s
\newcommand\covariancematrix\Sigma

\newcommand{\FmMu}{\mu_t(x_0)}
\newcommand{\FmSigma}{\boldsymbol{\sigma_t}(x_0)}
\newcommand{\FmSigmaTimeDerivative}{\frac{\partial\boldsymbol{\sigma_t}(x_0)}{\partial t}}

\newcommand\SigmaScalar\sigma

\usepackage{hyperref}
\usepackage{url}
\usepackage[utf8]{inputenc} 
\usepackage[T1]{fontenc}    
\usepackage{hyperref}       
\usepackage{url}            
\usepackage{booktabs}       
\usepackage{amsfonts}       
\usepackage{nicefrac}       
\usepackage{microtype}      
\usepackage{xcolor}         

\title{Edge-preserving noise for diffusion models}

\author[Jente Vandersanden, Sascha Holl, Xingchang Huang, Gurprit Singh]
{\parbox{\textwidth}{\centering Jente Vandersanden$^{1}$\orcid{0000-0002-9488-9327}, Sascha Holl$^{1}$\orcid{0009-0004-7245-7998},
Xingchang Huang$^{1}$\orcid{0000-0002-2769-8408},
Gurprit Singh$^{1,2}$\orcid{0000-0003-0970-5835},
        }
        \\
{\parbox{\textwidth}{\centering $^1$Max Planck Institute for Informatics, Germany\\
         $^2$Advanced Micro Devices, Germany
       }
}
}

\begin{document}

\teaser{
 \includegraphics[width=1.0\linewidth]{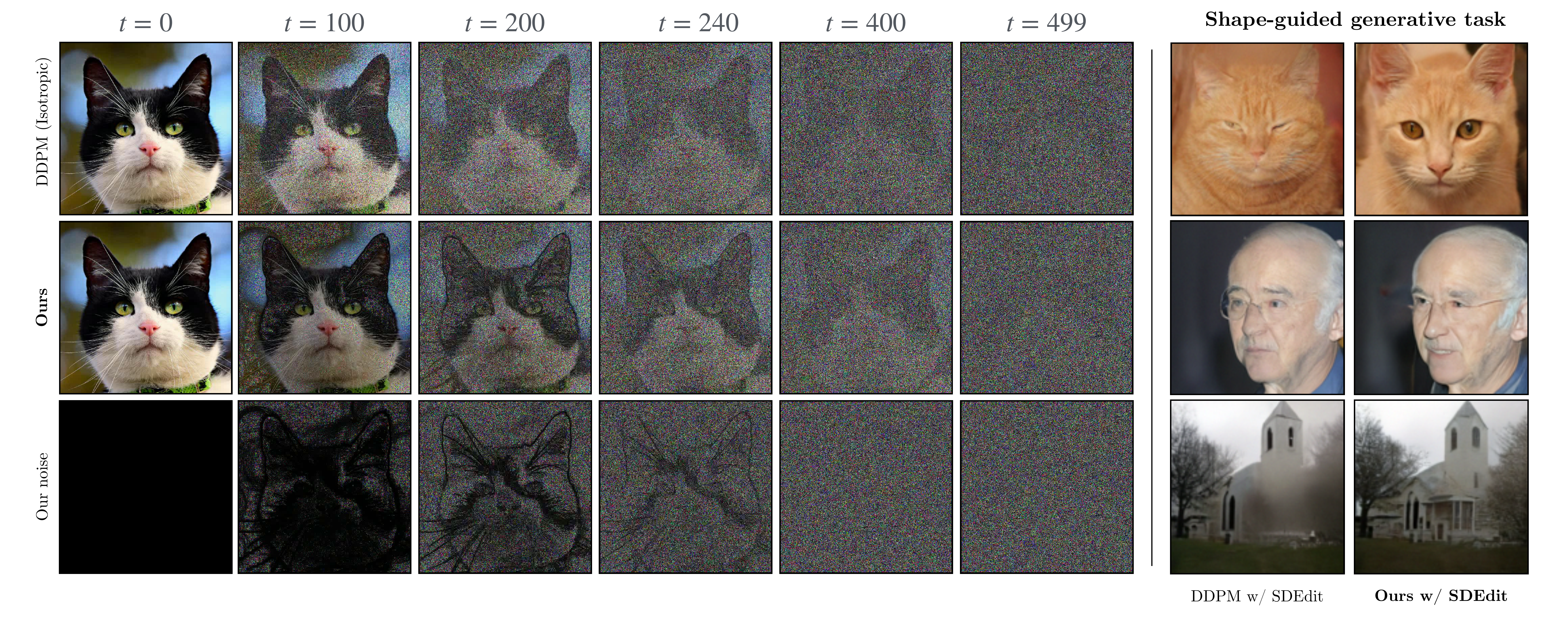}
 \centering
  \caption{A classic isotropic diffusion process (top row) is compared to our hybrid edge-aware diffusion process (middle row) on the left side. 
        We propose a hybrid noise schedule (bottom row) that smoothly transitions from anisotropic ($t \in [0;250[$) to isotropic noise ($t \in [250;499]$) . We use our edge-aware noise for training and inference.
        On the right, we compare both noise schemes applied to the SDEdit framework \cite{meng2021sdedit} for stroke-based image generation. Our model consistently outperforms DDPM's isotropic noise scheme, is more robust against visual artifacts and produces sharper outputs without missing structural details.}
\label{fig:teaser}
}

\maketitle
\begin{abstract}
Classical diffusion models typically rely on isotropic Gaussian noise, treating all regions uniformly and overlooking structural information important for high-quality generation. We introduce an edge-preserving diffusion process that generalizes isotropic models via a hybrid noise scheme with an edge-aware scheduler that smoothly transitions from edge-preserving to isotropic noise. This enables the model to capture fine structural details while generally maintaining global performance. We evaluate the impact of structure-aware noise in both diffusion and flow-matching frameworks, and show that existing isotropic models can be efficiently fine-tuned with edge-preserving noise, making our framework practical for adapting pre-trained systems. Beyond unconditional generation, our method particularly shows improvements in structure-guided tasks such as stroke-to-image synthesis, improving robustness and perceptual quality, as evidenced by consistent improvements across FID, KID, and CLIP-score.


\end{abstract}  
\section{Introduction}

Previous work on diffusion models mostly uses isotropic Gaussian noise to transform an unknown data distribution into a known distribution (e.g., normal distribution), from which samples can be efficiently drawn \cite{song2019generative, song2020score, ho2020denoising, kingma2021variational}. 
Due to the isotropic nature of the noise, all regions in the data samples $\Samplezero$ are uniformly corrupted, regardless of the underlying structural content, which is typically distributed in a non-isotropic manner. In the generative backward process, the model learns an isotropic denoising function, but in doing so, it ignores potentially valuable non-isotropic information in the data that it was trained on.
Denoising has been a central topic in image processing research \cite{elad2023image}. The seminal work by~\cite{perona1990scale} showed that accounting for image structure enables substantial gains in denoising performance.
Since generative diffusion models can also be seen as \emph{denoisers}, we ask ourselves: \textit{Can incorporating structural information from data samples improve the effectiveness of a generative diffusion process?} 

To explore our question, we introduce a new class of diffusion models that generalizes over existing isotropic models and explicitly learns a content-aware noise scheme. We call our noise scheme \emph{edge-preserving noise}.

To summarize, we make the following contributions: 
\begin{itemize}
    \item We introduce a novel class of content-aware diffusion models and show how it is a generalization of existing isotropic diffusion models (\cref{sec:ep_process_diff_models}). We also demonstrate that our noise framework can be applied in the more general setting of flow matching (\cref{sec:ep_process_flow_matching}).
    \item We run extensive qualitative and quantitative experiments across a variety of datasets to validate the positive impact of using edge-preserving noise over isotropic noise (\cref{sec:experiments} and \cref{sec:appendix_additional_results}).
    \item We analyze our model's generative process, and demonstrate that it converges more rapidly to sharper, less noisy predictions (\cref{fig:pred_x0s_comparison}). In addition, we conduct a frequency analysis, suggesting that our edge-preserving model better learns the low-to-mid frequencies of the target data (\cref{sec:frequency_analysis}).
    \item We observe consistent quantitive/qualitative improvements for unconditional image generation. In particular, our noise framework demonstrates strong potential for shape-guided generative tasks, showing greater robustness and significantly improved quality on these tasks (\cref{fig:sde_edit_comparison}). 
\end{itemize} 

\section{Related work}
Most existing diffusion-based generative models~\cite{sohl2015deep,song2019generative,song2020score,ho2020denoising} corrupt data samples by adding noise with the same variance across all pixels. 
Generative models tend to have the ability to produce more diverse and novel content when the noise variance is higher, whereas lower variance noise is better at preserving the underlying structure of the data. Various efforts have explored diffusion processes beyond those driven solely by isotropic noise.
\cite{rissanen2022generative} introduced an inverse heat dissipation model (IHDM), which applies isotropic Gaussian blurring to corrupt images, which they show is equivalent to introducing non-isotropic noise in the frequency domain.  One line of work ~\cite{bansal2022cold,daras2023soft} investigates arbitrary forward diffusion processes with mixed components such as blurring, noise, masking... \cite{hoogeboom2022blurring} propose a generalized form of heat dissipation and diffusion by combining isotropic noise and blurring.

Another line of work has explored non-isotropic forms of noise in diffusion models. \cite{dockhorn2022scorebased} proposed to use critically-damped Langevin diffusion where the data variable at any time is augmented with an additional "velocity" variable. Noise is only injected in the velocity variable. 
\cite{voleti2022score} performed a limited study on the impact of isotropic vs non-isotropic Gaussian noise for a score-based model. The idea behind non-isotropic Gaussian noise is to use noise with different variance across image pixels.
They use a non-diagonal covariance matrix to generate non-isotropic Gaussian noise, but their sample quality did not improve in comparison to the isotropic case.
\cite{yu2024constructing} developed this idea further and proposed a Gaussian noise model that adds noise with non-isotropic variance to pixels. The variance is chosen based on how much a pixel or region needs to be edited. They demonstrated a positive impact on editing tasks. More recently, \cite{huang2024blue} proposed a blue noise diffusion model (BNDM), using negatively correlated noise for enhanced visual quality and FID scores. While IHDM and BNDM also consider a form of non-isotropic noise, they do not explicitly account for structures present in the signal. 

Our definition of non-isotropy is inspired by the seminal work of \cite{perona1990scale} on anisotropic diffusion for edge-preserving image filtering (removing noise from images). We apply a non-isotropic variance to pixels in an edge-aware manner, meaning that we suppress noise on edges.

\section{Background}
\label{sec:background}

\subsection{Generative diffusion processes}
\label{sec:gaussian_diff_models}

A generative diffusion model consists of two processes: the forward process transforms data samples $\Samplezero$ into samples $\SampleT$ that are distributed according to a well-known prior distribution, such as a normal distribution~$\NormalDist(0,I)$. The corresponding backward process does exactly the opposite: it transforms samples $\SampleT$ into $\SamplezeroApprox$, distributed according to the target distribution $p_0(\mathbf{x})$. Sampling from this backward process involves predicting a vector quantity, interpretable as either noise or the gradient of the data distribution, which is precisely the task for which the generative diffusion model is trained. Previous works \cite{song2019generative, song2020score, ho2020denoising, kingma2021variational, rissanen2022generative, hoogeboom2022blurring} typically formulate the forward process as the following linear equation:

\begin{equation}
 \label{eq:gaussian_diff_linear_eq}
  \Samplet = \SignalCoeff_t \Samplezero + \SigmatScalar \Noise
\end{equation}
here, $\Samplet$ is the data sample diffused up to time $t$, $\Samplezero$ stands for the original data sample, $\Noise$ is a standard normal Gaussian noise, and the \textit{signal coefficient} $\SignalCoeff_t$ and \textit{noise coefficient} $\SigmatScalar$ determine the signal-to-noise ratio (SNR) ($\nicefrac{\SignalCoeff_t}{\SigmatScalar}$). The SNR refers to the proportion of signal retained relative to the amount of injected noise. Note that $\SignalCoeff_t$ and $\SigmatScalar$ are both scalars. Previous works have made several different choices for $\SignalCoeff_t$ and $\SigmatScalar$ respectively, leading to different variants, each with their own advantages and limitations. Typically, a \textit{noise schedule} $\beta_t$ is employed to govern the rate at which $\SignalCoeff_t$ and $\SigmatScalar$ vary over time \cite{ho2020denoising}. We define $\alpha_t = 1 - \beta_t$ and $\bar{\alpha_t} = \AlphaCumProd$, consistent with \cite{ho2020denoising}.

\subsection{Denoising probabilistic models}
 Following the probabilistic paradigm of \cite{ho2020denoising}, we would like to introduce the posterior probability distributions of the general diffusion process described by \cref{eq:gaussian_diff_linear_eq}. We will show the exact form that our forward and backward processes take in \cref{sec:hybrid_noise} and \cref{sec:ep_process_diff_models} respectively. For details and full derivations of the equations provided in this paragraph, we would like to refer to  the appendix of \cite{kingma2021variational}. The isotropic diffusion process formulated in~\cref{eq:gaussian_diff_linear_eq} has the following marginal distribution: 
\begin{equation}
 \label{eq:gaussian_diffusion_marginal}
  q(\Samplet|\Samplezero) = \NormalDist(\SignalCoeff \Samplezero, \SigmatScalar^2 \Identity)
\end{equation}
Moreover, it has the following Markovian transition rules: 
\begin{equation}
\label{eq:gaussian_diffusion_transition_probabilities}
  q(\Samplet|\Samples) = \NormalDist(\Alphatovers \Samples, \Sigmatovers^2 \Identity)
\end{equation}
with the forward posterior signal coefficient 
$
  \Alphatovers = \frac{\Alphat}{\Alphas}
$
and the forward posterior variance (or square of the noise coefficient)
%
  $\Sigmatovers^2 = \SigmatScalar^2 - \Alphatovers^2 \SigmasScalar^2$
%
and $0< s < t < T$. 
For a Gaussian diffusion process, given that $q(\Samples|\Samplet, \Samplezero) \propto q(\Samplet|\Samples) q(\Samples|\Samplezero)$, one can analytically derive a \textit{backward process} that is also Gaussian, and has the following marginal distribution:
\begin{equation}
 \label{eq:gaussian_reverse_marginal_dist}
  q(\Samples|\Samplet, \Samplezero) = \NormalDist(\Meanttos, \Sigmattos^2 \Identity).
\end{equation}
The backward posterior variance $\Sigmattos^2$ has the following form:
\begin{equation}
 \label{eq:sigma_t_to_s_sqrd}
    \Sigmattos^2 = \left(\frac{1}{\sigma_s^2} + \frac{\Alphatovers^2}{\sigma_{t|s}^2}\right)^{-1}
\end{equation}
and the backward posterior mean $\Meanttos$ is formulated as:
\begin{equation}
 \label{eq:mu_t_to_s}
    \Meanttos = \Sigmattos^2 \left(\frac{\Alphatovers}{\Sigmatovers^2} \Samplet + \frac{\Alphas}{\SigmasScalar^2} \Samplezero \right).
\end{equation}

Samples can be generated by simulating the reverse Gaussian process with the posteriors in \cref{eq:sigma_t_to_s_sqrd} and \cref{eq:mu_t_to_s}. A practical issue is that \cref{eq:mu_t_to_s} itself depends on the unknown $\Samplezero$, the sample we are trying to generate. To overcome this, one can instead approximate the analytic reverse process in which $\Samplezero$ is replaced by its approximator $\SamplezeroApprox$, learned by a deep neural network $\Model$. The network can learn to directly predict $\Samplezero$ given an $\Samplet$ (a sample with a level of noise that corresponds to time $t$), but previous work \cite{ho2020denoising} has shown that it is beneficial to instead optimize the network to learn the approximator $\NoiseApprox$. $\NoiseApprox$ predicts the unscaled Gaussian white noise that was injected at time $t$. $\SamplezeroApprox$ can then be obtained via \cref{eq:hat_x_0}, which follows from \cref{eq:gaussian_diff_linear_eq}.
\begin{equation}
 \label{eq:hat_x_0}
    \SamplezeroApprox = \frac{1}{\Alphat} \Samplet - \frac{\SigmatScalar}{\Alphat} \NoiseApprox
\end{equation}
%
\vspace{-4mm}

\subsection{Edge-preserving filters in image processing}
\label{sec:edge_preserving_filters}
In this work, we aim to choose $\SignalCoeff_t$ and $\SigmatScalar$ such that we obtain a diffusion process that injects noise in a content-aware manner. To do this, we are inspired by the field of image processing, where a classic and effective technique for denoising is edge-preserved filtering via \textit{anisotropic diffusion} \cite{weickert1998anisotropic}. To overcome the problem of destroying relevant structural information in the image when applying an isotropic filter, \cite{perona1990scale} instead propose an anisotropic diffusion process of the form: 
\begin{equation}
 \label{eq:perona_malik_aniso_diffusion}
    \Samplet = \Samplezero + \int_{0}^{t} \DiffusionCoeffTensorXs^{-1} \LaplaceSample_s \,ds
\end{equation}
where the \textit{diffusion coefficient} $\DiffusionCoeffTensorXs^{-1}$ has the following general form:
\begin{equation}
 \label{eq:perona_malik_diff_coeff}
    \DiffusionCoeffTensor = \PeronaMalikFilterTimeDependentReciprocal
\end{equation}
where $\ImageGradient$ is the gradient magnitude image, and $\EdgeSensitivityTimeVarying$ is the \textit{edge sensitivity}. Intuitively, in the regions of the image where the gradient response is high (on edges), the diffusion coefficient will be smaller, and therefore the signal gets less distorted there. The edge sensitivity $\EdgeSensitivityTimeVarying$ determines how sensitive the diffusion coefficient is to the image gradient response. Note that in the original formulation of \cite{perona1990scale}, $\EdgeSensitivityTimeVarying$ is introduced as a constant parameter. We define the diffusion coefficient with the same constant $\EdgeSensitivity$ as:

\begin{equation}
 \label{eq:diff_coeff_non_time_dependent}
    \DiffusionCoeffTensorNonTimeDependent = \DiffusionCoeffTensorZero = \PeronaMalikFilterReciprocal
\end{equation}
but we found that a time-varying edge sensitivity works better in our generative setting (more details in \cref{sec:time_varying_edge_sensitivity,sec:ablation}).

Inspired by the anisotropic diffusion coefficient presented in \cref{eq:perona_malik_diff_coeff}, we aim to design a \textit{linear diffusion process} that incorporates edge-preserving noise. Our hope is that by doing this, the generative diffusion model will better learn the underlying geometrical structures of the target distribution, leading to a more effective generative denoising process. To obtain our content-aware linear diffusion process, we apply the idea of edge-preserved filtering to the noise term of \cref{eq:gaussian_diff_linear_eq}. We cannot directly use \cite{perona1990scale}'s formulation because their time-dependent diffusion coefficient makes the process nonlinear (in \cref{eq:perona_malik_aniso_diffusion}, the diffusion coefficient $\DiffusionCoeffTensor$ depends on $\Sample_s$). Instead, we make the coefficient depend only on $\Samplezero$:

\begin{equation}
 \label{eq:our_forward_process}
    \Samplet = \SignalCoeff_t \Samplezero + \SigmaOursGeneral \Noise
\end{equation}
Where $\NoiseCoeffNumerator$ is the noise coefficient's numerator and can be chosen as desired. To investigate the mere impact of non-isotropic edge-preserving noise on the generative diffusion process, we chose our parameters $\SignalCoeff_t = \SqrtAlphatBar$ and $\NoiseCoeffNumerator = \SqrtOneMinAlphatBar$ such that it closely matches the well-studied forward process of \cite{ho2020denoising}, but nothing prevents us from making different choices for $\SignalCoeff_t$ and $\NoiseCoeffNumerator$. Note that the noise coefficient in \cref{eq:gaussian_diff_linear_eq} becomes a tensor $\SigmatTensor$ instead of a scalar $\SigmatScalar$ for our process. Intuitively, we preserve edges by reducing noise based on the edges in the \textit{original} image. 

\section{An edge-preserving generative process}
\label{sec:edge_preserving_noise_our_diffusion_model}

\subsection{Forward process with hybrid noise scheme}
\label{sec:hybrid_noise}

The forward \emph{edge-preserving} process described in~\cref{eq:our_forward_process} in its pure form is not very meaningful in a generative setting. This is because if the edges are preserved all the way up to time $t=T$, we end up with a rather complex \emph{prior} distribution $p_T(x)$ that we cannot efficiently take samples from. Instead, we would like to end up with a well-known  distribution at time $t=T$, such as the standard normal distribution. To achieve this, we instead consider the following hybrid forward process: 

\begin{equation}
 \label{eq:our_forward_process_time_dependent}
    \Samplet = \SignalCoeff_t \Samplezero + \SigmaOursTimeVaryingGeneral \Noise
\end{equation}

The function $\TransitionFn$ now appearing in the denominator of the diffusion coefficient is the \textit{transition function}. When $\TransitionFn < 1$, we obtain edge-preserving noise (the edge-preservation is strongest when $\TransitionFn \approx 0$). The turning point where $\TransitionFn = 1$ is called the \textit{transition point} $\TransitionPt$. At the transition point, we switch over to isotropic noise with scalar noise coefficient 
$\SigmatScalar = \NoiseCoeffNumerator$ (note that we chose $\SignalCoeff_t = \SqrtAlphatBar$ and $\NoiseCoeffNumerator = \SqrtOneMinAlphatBar$). 

This approach allows us to flexibly design noise schedulers that start off with edge-preserving noise and towards the end of the forward process fall back to an isotropic diffusion coefficient. Practically, one can choose any function for $\TransitionFn$, as long as it maps to $[0;1]$ and $\TransitionFn = 1$ for $t$ in proximity to $T$. We performed an ablation for different transition functions in \cref{sec:ablation}. 

Observe how our diffusion process generalizes over existing isotropic processes: by setting $\TransitionFn = 1$ constant, we simply obtain an isotropic process with signal coefficient $\SignalCoeff_t$ and noise coefficient $\SigmatScalar = \NoiseCoeffNumerator$. Choosing any other non-constant function for $\TransitionFn$ leads to a hybrid diffusion process that consists of an edge-preserving stage and an isotropic stage (starting at $\TransitionFn=1$). 


\begin{figure}[htb]
  \centering
  \includegraphics[width=1.0\linewidth]{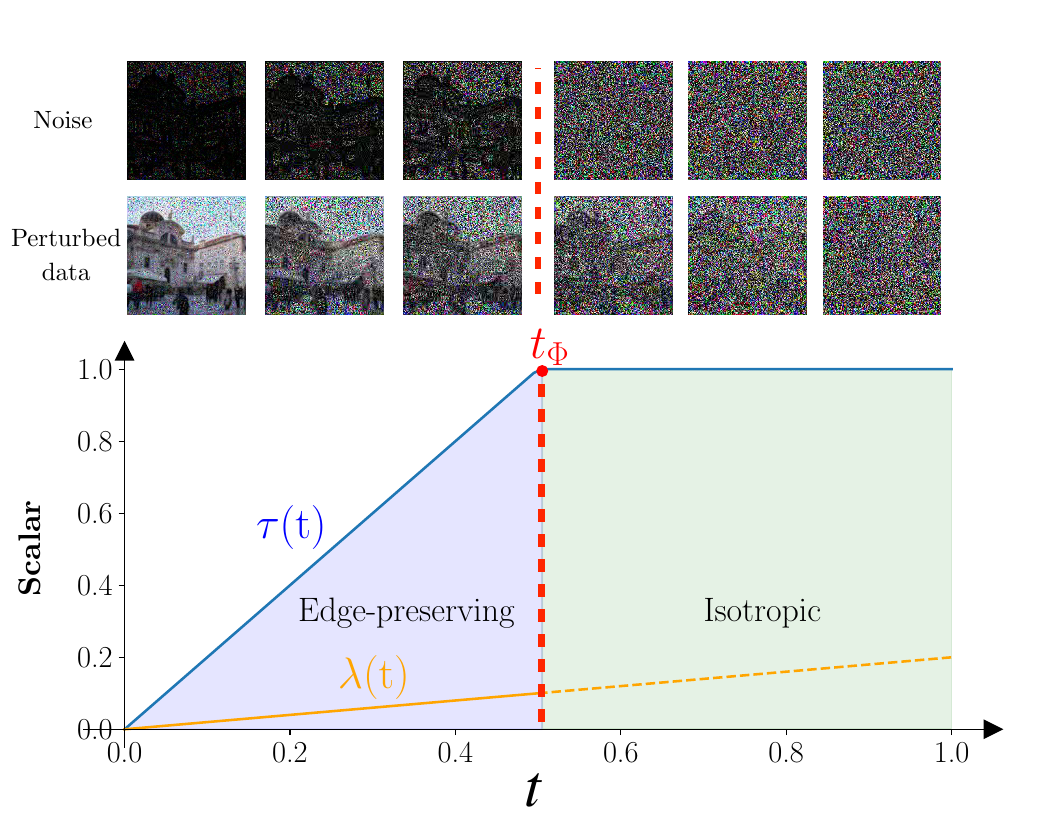}
  \caption{Illustration of our hybrid diffusion process and its hyperparameters. All parameters are explained in \cref{sec:hybrid_noise,sec:time_varying_edge_sensitivity}.}
  \label{fig:transition_fns_notext}
\end{figure}

\subsection{Time-varying edge sensitivity $\EdgeSensitivityTimeVarying$}
\label{sec:time_varying_edge_sensitivity}
The edge sensitivity parameter $\EdgeSensitivity$ controls the level of detail preserved along image edges. Very low values of (e.g. $\EdgeSensitivity=1e-5$) will retain almost all fine details. The more we increase $\EdgeSensitivity$, the less details will be preserved. When $\EdgeSensitivity$ becomes very high (e.g. $\EdgeSensitivity=1$), the process becomes nearly isotropic. Our ablation study (\cref{sec:ablation}) explores impact of this parameter in more detail. We found that constant $\EdgeSensitivity$-values harm sample quality: too low values results in unrealistic, "cartoonish" images, while too high values  diminish the effectiveness of the edge-preserving diffusion model, making the model behave almost like an isotropic process.

To address this, we instead consider a time-varying edge sensitivity $\EdgeSensitivityTimeVarying$. We set an interval $[\EdgeSensitivityMin; \EdgeSensitivityMax]$ that bounds the possible values for the time-varying edge sensitivity. The function that governs $\EdgeSensitivityTimeVarying$ within this interval can in theory again be chosen freely. We have so far experimented with a linear function and a sigmoid function. We experienced that a linear function for $\EdgeSensitivityTimeVarying$ resulted in higher sample quality and therefore used this function for our experiments. Additionally, we have attempted to optimize the interval $[\EdgeSensitivityMin; \EdgeSensitivityMax]$, but this led to unstable behaviour.

\subsection{Edge-Aware Generative Process in Diffusion Models}
\label{sec:ep_process_diff_models}

Given the forward hybrid diffusion process introduced in \cref{sec:hybrid_noise}, we construct the corresponding generative backward process within the denoising diffusion framework (for the edge-preserving flow-matching variant, see \cref{sec:ep_process_flow_matching}). Specifically, we derive explicit expressions for the posterior mean $\Meanttos$ and variance $\SigmattosTensor^2$ of the backward process by substituting our chosen signal coefficient $\SignalCoeff_t$ and variance $\SigmatTensor^2$ into \cref{eq:mu_t_to_s} and \cref{eq:sigma_t_to_s_sqrd}.Recall that we chose $\SigmatTensor^2$ to be a tensor, which is why the backward posterior variance $\SigmattosTensor^2$ is again a tensor, contrary to isotropic diffusion processes considered in previous works. Regardless, we can use the same equations and the algebra still works. 

We first introduce an auxiliary variable $\AuxiliaryOne$, which represents the variance of our forward process at a given time $t$. This is simply the square of our choice for the noise coefficient $\SigmatTensor$ formulated in ~\cref{eq:our_forward_process_time_dependent}:
\begin{equation}
 \label{eq:auxiliary_one}
    \begin{split}
       \AuxiliaryOne&=(1-\AlphatBar)\cdot\Bigl[(1-\TransitionFn)^2\DiffusionCoeffTensorXZero^2\\
       &\;\;\;\;\;\;\;\;\;\;\;\;+2\;(1 -\TransitionFn)\DiffusionCoeffTensorXZero\TransitionFn+\TransitionFn^2\Bigr]^{-1}
    \end{split}
\end{equation}
Here $\AlphatBar$ has the same meaning as earlier described in  \cref{sec:gaussian_diff_models}. We now have the backward posterior variance $\SigmattosTensor^2$:
\begin{equation}
 \label{eq:our_backward_posterior_variance}
    \SigmattosTensor^2 = \OurBackwardPosteriorVariance
\end{equation}
and the backward posterior mean $\Meanttos$: 
\begin{equation}
 \label{eq:our_backward_posterior_mean}
    \Meanttos = \OurBackwardPosteriorMean
\end{equation}
Given \cref{eq:our_backward_posterior_variance} and \cref{eq:our_backward_posterior_mean}, the only unknown preventing us from simulating the Gaussian backward process is $\Samplezero$. Note that $\Samplezero$ in our case depends on a non-isotropic noise. 
Therefore, we cannot just use an isotropic approximator $\NoiseApprox$ for the isotropic noise $\Noise$ to predict $\SamplezeroApprox$ via \cref{eq:hat_x_0}. 
Instead, we need a model that can predict the non-isotropic noise $\SigmatTensor \Noise$ . 

\begin{figure*}[tbp]
    \centering

\newcommand{\PlotSingleImage}[1]{%
        \begin{scope}
            \clip (0,0) -- (2.5,0) -- (2.5,2.5) -- (0,2.5) -- cycle;
            \path[fill overzoom image=figures/pred_x0s_comparison/#1] (0,0) rectangle (2.5cm,2.5cm);
        \end{scope}
        \draw (0,0) -- (2.5,0) -- (2.5,2.5) -- (0,2.5) -- cycle;
        
}

\newcommand\scalevalueBigger{0.7}    

\small
\hspace*{-4mm}

    

\begin{tabular}{c@{\;} c@{\;} c@{\;} c@{\;} c@{\;} c@{\;} c@{\;} c@{\;} c@{\;} c@{\;}}
~ &
$t=499$ & 
$t=460$ &
$t=400$ &
$t=340$ &
$t=280$ &
$t=250$ &
$t=180$ &
$t=90$ &
$t=0$ 
\\
\rotatebox{90}{\hspace{0.45cm} \scriptsize Isotropic}
&
\begin{tikzpicture}[scale=\scalevalueBigger]
\PlotSingleImage{ddpm/0.png}
\end{tikzpicture}
&
\begin{tikzpicture}[scale=\scalevalueBigger]
\PlotSingleImage{ddpm/1.png}
\end{tikzpicture}
&
\begin{tikzpicture}[scale=\scalevalueBigger]
\PlotSingleImage{ddpm/2_new_2.png}
\end{tikzpicture}
&
\begin{tikzpicture}[scale=\scalevalueBigger]
\PlotSingleImage{ddpm/3.png}
\end{tikzpicture}
&
\begin{tikzpicture}[scale=\scalevalueBigger]
\PlotSingleImage{ddpm/4.png}
\end{tikzpicture}
&
\begin{tikzpicture}[scale=\scalevalueBigger]
\PlotSingleImage{ddpm/5.png}
\end{tikzpicture}
&
\begin{tikzpicture}[scale=\scalevalueBigger]
\PlotSingleImage{ddpm/6.png}
\end{tikzpicture}
&
\begin{tikzpicture}[scale=\scalevalueBigger]
\PlotSingleImage{ddpm/7.png}
\end{tikzpicture}
&
\begin{tikzpicture}[scale=\scalevalueBigger]
\PlotSingleImage{ddpm/8.png}
\end{tikzpicture}
\\
\rotatebox{90}{\hspace{0.65cm} \scriptsize Ours}
&
\begin{tikzpicture}[scale=\scalevalueBigger]
\PlotSingleImage{ours/0.png}
\end{tikzpicture}
&
\begin{tikzpicture}[scale=\scalevalueBigger]
\PlotSingleImage{ours/1.png}
\end{tikzpicture}
&
\begin{tikzpicture}[scale=\scalevalueBigger]
\PlotSingleImage{ours/2_new_2.png}
\end{tikzpicture}
&
\begin{tikzpicture}[scale=\scalevalueBigger]
\PlotSingleImage{ours/3.png}
\end{tikzpicture}
&
\begin{tikzpicture}[scale=\scalevalueBigger]
\PlotSingleImage{ours/4.png}
\end{tikzpicture}
&
\begin{tikzpicture}[scale=\scalevalueBigger]
\PlotSingleImage{ours/5.png}
\end{tikzpicture}
&
\begin{tikzpicture}[scale=\scalevalueBigger]
\PlotSingleImage{ours/6.png}
\end{tikzpicture}
&
\begin{tikzpicture}[scale=\scalevalueBigger]
\PlotSingleImage{ours/7.png}
\end{tikzpicture}
&
\begin{tikzpicture}[scale=\scalevalueBigger]
\PlotSingleImage{ours/8.png}
\end{tikzpicture}
%

    

\end{tabular} 
    \vspace{-0.5mm}
    \caption{We visually compare the impact of our edge-preserving noise on the generative process. In each column, we show predictions $\SamplezeroApprox$ at selected  time steps. Our method converges significantly faster to a sharper and less noisy image than its isotropic counterpart. This is evident by the earlier emergence (from $t=400$) of structural details like the pattern on the cat's head, eyes, and whiskers with our approach.}
\label{fig:pred_x0s_comparison}
\end{figure*}


We introduce the loss function that trains such an approximator:
\begin{equation}
 \label{eq:loss_fn}
    \Loss = \OurLossFn.
\end{equation}
It is very similar to the simplified loss function derived in DDPM \cite{ho2020denoising}, \textbf{with the difference that our model explicitly learns to predict the non-isotropic edge-preserving noise ($\SigmatTensor\Noise$)}. 
Note that we apply no weighting to our loss function. In~\cref{sec:negative_log_likelihood}, we discuss that this is a heuristic, and we also show how our loss formulation can be derived from a negative log-likelihood perspective, with the accurate theoretically-founded weighting.

$\Model$ stands for the time-conditioned U-Net used to approximate the time-varying noise function. The visual difference between the backward process of an isotropic diffusion model (DDPM) and ours is shown in \cref{fig:pred_x0s_comparison}. Our formulation introduces a negligible overhead. The only additional computation that needs to be performed is the image gradient $\StartImageGradient$, which can be done very efficiently on modern GPUs. We have not noticed any significant difference in training time between vanilla DDPM and our method.

\begin{figure*}[tbp]
    \centering

\newcommand{\PlotSingleImage}[1]{%
        \begin{scope}
            \clip (0,0) -- (2.5,0) -- (2.5,2.5) -- (0,2.5) -- cycle;
            \path[fill overzoom image=figures/#1] (0,0) rectangle (2.5cm,2.5cm);
        \end{scope}
        \draw (0,0) -- (2.5,0) -- (2.5,2.5) -- (0,2.5) -- cycle;
        
}

\newcommand{\PlotSingleImageWithLine}[1]{%
        \begin{scope}
            \clip (0,0) -- (2.5,0) -- (2.5,2.5) -- (0,2.5) -- cycle;
            \path[fill overzoom image=figures/#1] (0,0) rectangle (2.5cm,2.5cm);
        \end{scope}
        \draw (0,0) -- (2.5,0) -- (2.5,2.5) -- (0,2.5) -- cycle;
        \draw[dashed] (0, -0.13) -- (2.5, -0.13);
}

\newcommand{\TwoColumnFigure}[2]{%
    \begin{tabular}{c@{\;}c@{}}
        \hspace*{-2.5mm}
        \begin{tikzpicture}[scale=0.563]
            \PlotSingleImage{#1}
        \end{tikzpicture}
         & 
         \begin{tikzpicture}[scale=0.563]
            \PlotSingleImage{#2}
        \end{tikzpicture}
    \end{tabular}%
}
\newcommand\scalevalue{0.95}    
\newcommand\smallerscale{0.73}    

\hspace*{-5mm}
\begin{tabular}{c@{\;}c@{}}
\footnotesize
\begin{tabular}{c@{\;}c@{\;}c@{\;}c@{\;}c@{}}


&
\begin{tikzpicture}[scale=\scalevalue]
\PlotSingleImage{sde_edit_comparison/celeba/stroke_painting_from_image.png}
\end{tikzpicture}
&
\begin{tikzpicture}[scale=\scalevalue]
\PlotSingleImage{sde_edit_comparison/celeba/bndm_generated.png}
\end{tikzpicture}
&
\begin{tikzpicture}[scale=\scalevalue]
\PlotSingleImage{sde_edit_comparison/celeba/ddpm_generated.png}
\end{tikzpicture}
&
\begin{tikzpicture}[scale=\scalevalue]
\PlotSingleImage{sde_edit_comparison/celeba/ours_generated.png}
\end{tikzpicture}
\\
&
\begin{tikzpicture}[scale=\scalevalue]
\PlotSingleImage{sde_edit_comparison/church/extras/painting_605.png}
\end{tikzpicture}
&
\begin{tikzpicture}[scale=\scalevalue]
\PlotSingleImage{sde_edit_comparison/church/extras/bndm_605.png}
\end{tikzpicture}
&
\begin{tikzpicture}[scale=\scalevalue]
\PlotSingleImage{sde_edit_comparison/church/extras/ddpm_605.png}
\end{tikzpicture}
&
\begin{tikzpicture}[scale=\scalevalue]
\PlotSingleImage{sde_edit_comparison/church/extras/ours_605.png}
\end{tikzpicture}
\\
&
\begin{tikzpicture}[scale=\scalevalue]
\PlotSingleImage{sde_edit_comparison/cat/stroke_painting_from_image.png}
\end{tikzpicture}
&
\begin{tikzpicture}[scale=\scalevalue]
\PlotSingleImage{sde_edit_comparison/cat/bndm_generated.png}
\end{tikzpicture}
&
\begin{tikzpicture}[scale=\scalevalue]
\PlotSingleImage{sde_edit_comparison/cat/ddpm_generated.png}
\end{tikzpicture}
&
\begin{tikzpicture}[scale=\scalevalue]
\PlotSingleImage{sde_edit_comparison/cat/ours_generated.png}
\end{tikzpicture}
\\
& Synthetic & Blue noise & Isotropic noise  & \textbf{Ours} \\ & stroke painting &  &  & 
\\
\end{tabular}
&
\hspace{0.025cm}
\vline width 0.7 pt
\hspace{1mm}

\parbox[c]{5.8cm}{
\tiny
\centering

\newcolumntype{C}[1]{>{\centering\arraybackslash}p{#1}}

\begin{tabular}{|C{1.2cm}|C{0.9cm}|C{0.9cm}|C{0.9cm}|C{0.9cm}|}

\toprule
FID ($\downarrow$) & Blue noise & Isotropic & \textbf{Ours} \\
\midrule
CelebA($128^2$) & 68.0  & 45.80  & \textbf{39.08} \\
Church($128^2$) & 93.81 & 72.54  & \textbf{56.14} \\
Cat($128^2$) & 51.05 & 27.61 & \textbf{23.50} \\
\bottomrule
\end{tabular} 
\\

\begin{tabular}{c@{\;}c@{\;}c@{}}
\scriptsize
\begin{tikzpicture}[scale=\smallerscale]
\PlotSingleImage{user_created_stroke_guide/celeba/stroke_painting.png}
\end{tikzpicture}
&
\begin{tikzpicture}[scale=\smallerscale]
\PlotSingleImage{user_created_stroke_guide/celeba/ddpm.png}
\end{tikzpicture}
&
\begin{tikzpicture}[scale=\smallerscale]
\PlotSingleImage{user_created_stroke_guide/celeba/ours.png}
\end{tikzpicture}
\\
\begin{tikzpicture}[scale=\smallerscale]
\PlotSingleImage{user_created_stroke_guide/cat/stroke_painting.png}
\end{tikzpicture}
&
\begin{tikzpicture}[scale=\smallerscale]
\PlotSingleImage{user_created_stroke_guide/cat/sdedit_ddpm.png}
\end{tikzpicture}
&
\begin{tikzpicture}[scale=\smallerscale]
\PlotSingleImage{user_created_stroke_guide/cat/ours_sdedit.png}
\end{tikzpicture}
\\
\begin{tikzpicture}[scale=\smallerscale]
\PlotSingleImage{user_created_stroke_guide/church/stroke_painting.png}
\end{tikzpicture}
&
\begin{tikzpicture}[scale=\smallerscale]
\PlotSingleImage{user_created_stroke_guide/church/church_sdedit_ddpm.png}
\end{tikzpicture}
&
\begin{tikzpicture}[scale=\smallerscale]
\PlotSingleImage{user_created_stroke_guide/church/church_sdedit_ours.png}
\end{tikzpicture}
\\
 Human  & Isotropic noise & \textbf{Ours} \\ 
 painting &  &
\end{tabular}
}

\end{tabular} 
    \caption{\textbf{Left:} Impact of different types of noise to the SDEdit framework \cite{meng2021sdedit} for shape-guided generation. The leftmost column displays the stroke-based guide (created via k-means clustering applied to an image), with the other three columns showing the model outputs. Overall, using our noise franework results in sharper details and less distortions compared to other noises, leading to a better visual and quantitative performance. The corresponding FID scores are shown in the top right column. 
    \textbf{Right:} Our noise also works effectively with human-drawn paintings as shape guides, showing particularly precise adherence to details, such as the orange patches on the cat's fur.}
     \label{fig:sde_edit_comparison}
     \vspace{-4mm}
\end{figure*}

\subsection{Edge-Aware Generative Process in Flow Matching}
\label{sec:ep_process_flow_matching}
The general framework of flow matching allows users to design probability paths that on their turn will correspond to some probability flow vector field. Our goal is to construct a such path that leads to flows that are aware of the geometric structures in the target dataset. Motivated by its simple formulation and the impressive results \cite{lipman2022flow} achieved with it, we choose to build upon the optimal transport variant of flow matching (OT-FM). Theorem 3 derived by \cite{lipman2022flow} provides an elegant and flexible design framework for probability flows, where the user only has to specify differentiable functions $\FmMu$ and $\FmSigma$. These functions correspond to the signal coefficient $\SignalCoeff_t$ and noise coefficient $\SigmatScalar$ that we introduced in \cref{sec:background}. The OT-FM formulation chooses $\FmMu = t$ and $\FmSigma = 1 - t$. Similar to what we did for isotropic diffusion  (\cref{sec:edge_preserving_noise_our_diffusion_model,sec:ep_process_diff_models}), we make this probability path "edge-preserving" by leaving $\FmMu$ unchanged, and only operate on $\FmSigma$:

\vspace{-5mm}
\begin{align}
    \FmSigma = \frac{1-t}{(1 - \TransitionFn) \DiffusionCoeffTensorXZero + \TransitionFn}
\end{align}

To use this formulation in the framework of flow matching, we also need to find its corresponding time derivative: 
\begin{align}
    \FmSigmaTimeDerivative = \frac{g f' - fg'}{g^2}
\end{align}
Which follows from the quotient rule for derivatives, where $f$ is the numerator of $\FmSigma$ , $g$ is the denominator of $\FmSigma$, and $f'$ and $g'$ are its respective time derivatives: $f'=-1$, and $g'$ has the following form: 


\begin{align}
    g' &= \TransitionFnTimeDerivative + 
            (1-\TransitionFn)
            \biggl(
                \frac{\EdgeSensitivityTimeVarying 
                      - \EdgeSensitivityTimeVaryingTimeDerivative}
                     {2 \EdgeSensitivityTimeVarying^2 
                      \DiffusionCoeffTensorXZero}
            \biggr) -\TransitionFnTimeDerivative
            \DiffusionCoeffTensorXZero
\end{align}

However, note that because $\FmSigma$ has the requirement to be differentiable, $f$ and $g$ should also be differentiable. For $f$ there are trivially no issues, but $g$ contains two nested functions $\TransitionFn$ and $\EdgeSensitivityTimeVarying$, which are not necessarily differentiable. For example, the linear choice we made for $\TransitionFn$ (see \cref{fig:transition_fns_notext}) is only piecewise differentiable. We experimentally found that using this formulation leads to an unstable optimization objective, preventing the model from convergence. To overcome this, we tried to simplify our choice for $\FmSigma$. In particular, we removed the time dependency on the edge sensitivity $\EdgeSensitivity$, and changed $\TransitionFn$ to be piecewise constant instead of a piecewise linear function:
\vspace{-3mm}
\begin{align}
\TransitionFn =
    \begin{cases} 
      0 & t\leq \TransitionPt \\
      1 & t > \TransitionPt
   \end{cases}
\end{align}

This results to the time-derivative $g'$ of the denominator of $\FmSigma$ now becoming: 
\begin{align}
    g' = \frac{\partial\DiffusionCoeffTensorNonTimeDependent}{\partial t} - \DiffusionCoeffTensorNonTimeDependent\TransitionFnTimeDerivative  + \TransitionFnTimeDerivative = 0
\end{align}

This is the case because $\EdgeSensitivity$ no longer depends on time $t$ (and therefore $\DiffusionCoeffTensor=: \DiffusionCoeffTensorNonTimeDependent$, see \cref{eq:diff_coeff_non_time_dependent}) and the time derivative of $\TransitionFn$ is now $\TransitionFnTimeDerivative = 0$ over the whole domain. As a consequence, $\FmSigmaTimeDerivative$ now becomes:

\begin{equation}
\begin{aligned}
\FmSigmaTimeDerivative 
    = \frac{f'}{g} = \frac{-1}{
        (1 - \TransitionFn)
        \DiffusionCoeffTensorNonTimeDependent
        + \TransitionFn
    }
\end{aligned}
\end{equation}

With this slightly simplified formulation, we experienced a much more stable training objective. This is the final formulation that we used to generate the results in \cref{fig:comparison_fm}.

\section{Experiments}
\label{sec:experiments}

\subsection{Implementation details}
We provide the implementation details for our experiments in  \autoref{sec:exp_implementation_details}. Please also find our analysis on the model's capacity of modeling different frequency bands in \autoref{sec:frequency_analysis}.

\subsection{Unconditional image generation}
\begin{figure*}[tbp]
    \centering
    \includegraphics[scale=0.25]{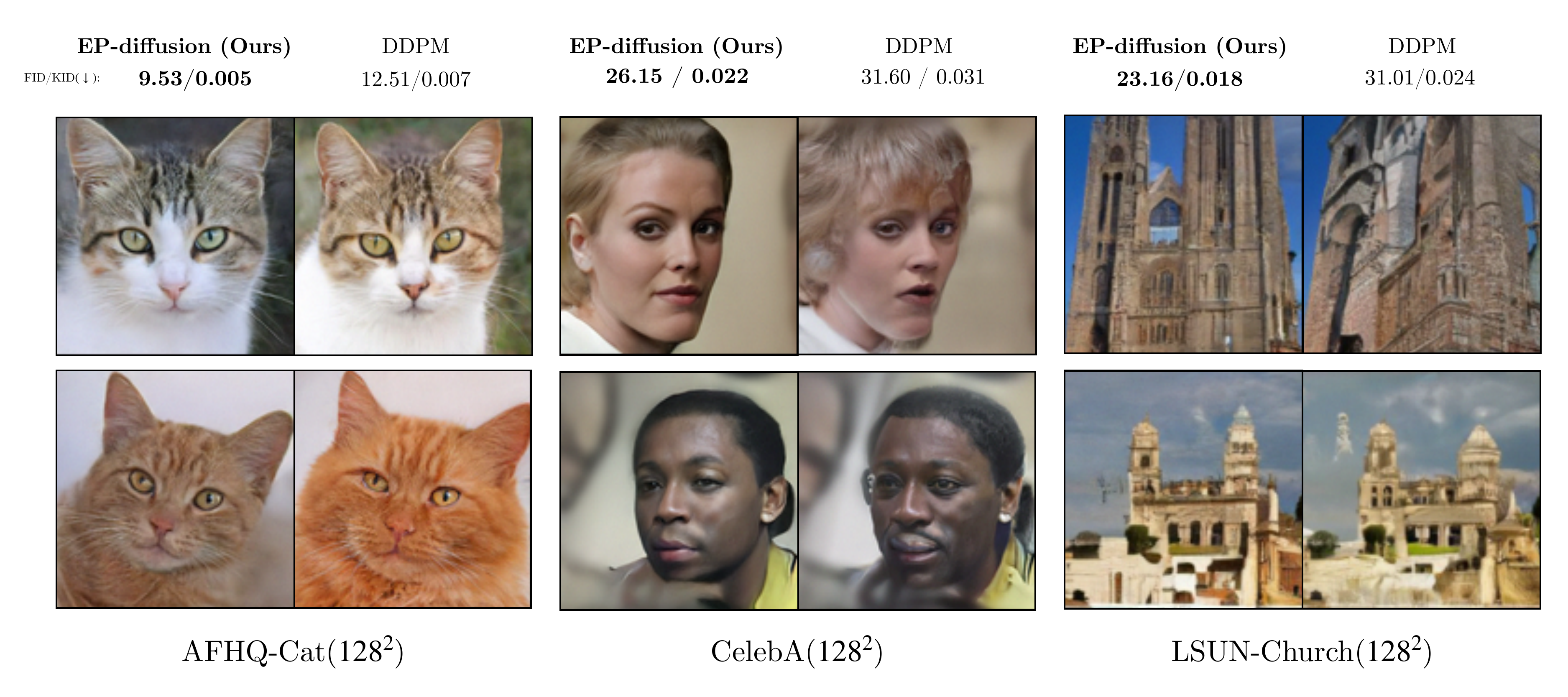}
    \caption{ 
Qualitative and quantitative comparison between edge-preserving noise vs. isotropic noise applied to the diffusion framework. Unconditional samples generated using our proposed edge-preserving noise model (\cref{sec:ep_process_diff_models}) are presented on the left side of each column and samples from the corresponding isotropic noise model (DDPM \cite{ho2020denoising}) are presented on the right side of each column. While qualitative differences can be subtle, the quantitative metrics (FID/KID, lower is better) reported above each column indicate that the edge-preserving noise model enhances the generative process. Additional results are provided in \cref{sec:appendix_additional_results}.}
    \label{fig:comparison}
\end{figure*}

\begin{figure*}[tbp]
    \centering
    \includegraphics[scale=0.25]{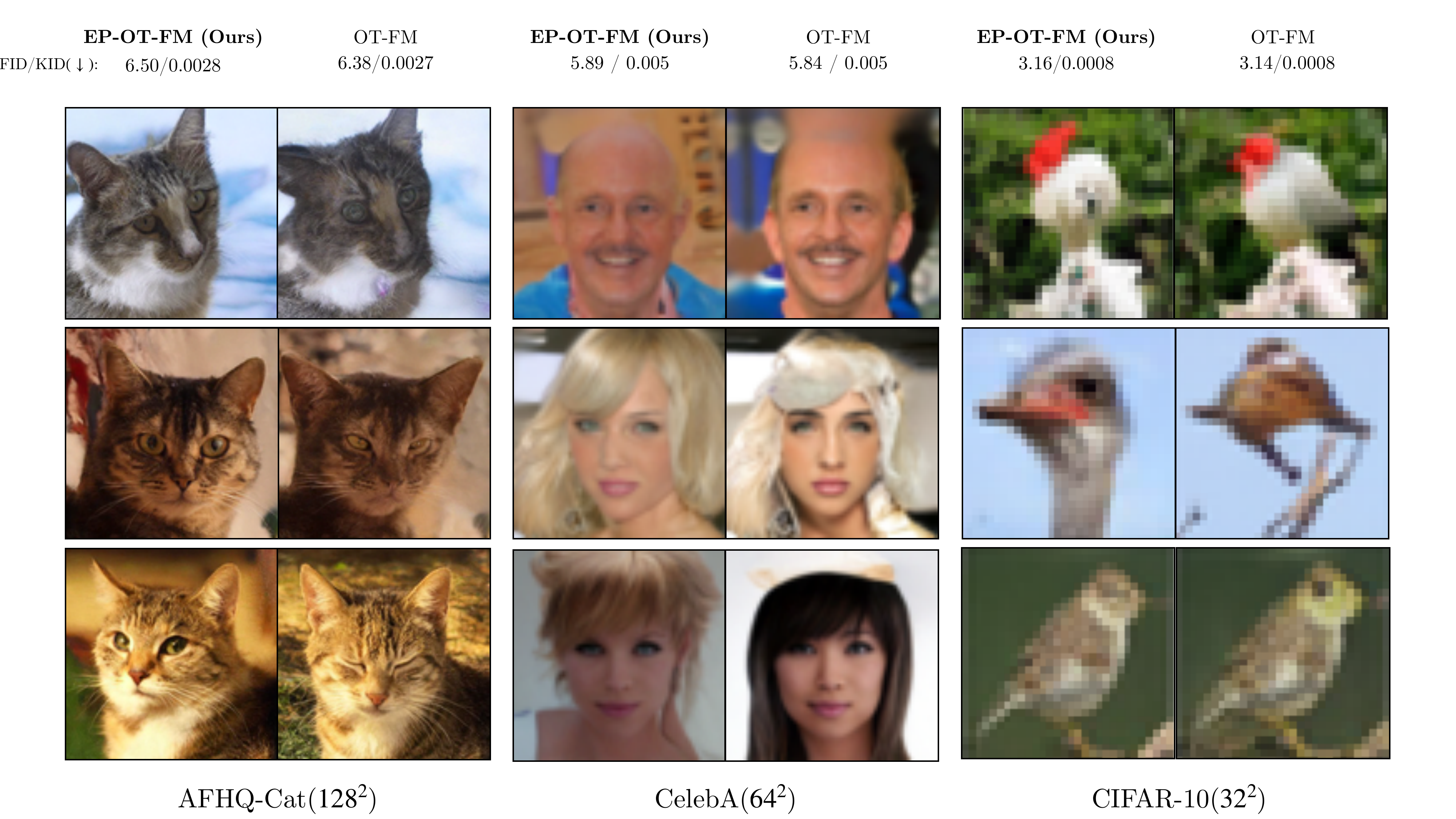}
    \vspace{2mm}
    \caption{Qualitative and quantitative comparison between edge-preserving noise vs. isotropic noise applied to the Flow Matching framework. Unconditional samples for \textbf{our edge-preserving variant (EP-OT-FM)} are displayed on the left side of each column, and the standard Optimal Transport Flow Matching (OT-FM) \cite{lipman2022flow} are displayed on the right. While FID and KID scores (lower is better) are closely matching, we observe that in the majority of time, our method delivers visual improvements in samples generated with the same seed. The results of our user study (\cref{tab:user_study_fm}) suggest the same.
    }
    \label{fig:comparison_fm}
\end{figure*}

We evaluate unconditional generation with edge-preserving noise in the diffusion framework, for which results are presented in \cref{fig:comparison} and \cref{sec:appendix_additional_results}). Across datasets, we found that edge-preserving (non-isotropic) noise brings improvements compared to isotropic noise \cite{ho2020denoising}. While qualitative gains over DDPM can be subtle for unconditional generation, we observed minor quantitative improvements, and found our method reduces visual artifacts and shows to be particularly effective in structure-guided generation (\cref{sec:stroke_guided_img_gen}, \cref{fig:sde_edit_comparison}).
To show that our noise scheduler also works in practice within the framework of flow matching, we compare OT-FM \cite{lipman2022flow} against an edge-preserving variant (EP-OT-FM) (see \cref{sec:ep_process_flow_matching} for details). Although FID/KID are similar, results in \cref{fig:comparison_fm} show consistent visual improvements. To validate this, we conducted an additional user study involving 30 participants. Each participant was presented with 10 pairs of generated images (for the AFHQ-Cat, CelebA, and CIFAR-10 datasets), with each pair consisting of one image generated by EP-Flow matching (ours) and one by OT-Flow matching. Participants were asked to rate each image on a 1–5 scale based on perceived quality (1 is the worst score, 5 is the best score).
The results show that EP-OT-FM was consistently preferred over OT-FM in terms of perceived sample quality (see \cref{tab:user_study_fm}).

\subsection{Edge-preserving noise in the latent space} 
We also briefly tested edge-preserving noise in the latent space, with results shown in \cref{tab:appendix_quantitative} and \cref{fig:appendix_cat_res512}. We would like to clarify that it makes sense to do this, given that in the latent space, a lot of geometric structure and shape of the original image is actually preserved (see \cref{fig:latents_visualization}).


\begin{table}[t]
\small
\centering
\caption{
Quantitative FID score comparison on latent-space diffusion \cite{rombach2022high} between DDPM~\cite{ho2020denoising} and our edge-preserving model.
}

\begin{tabular}{|c|c|c|}
\toprule
Unconditional FID ($\downarrow$) & AFHQ-Cat($512^2$, latent) \\
\midrule
DDPM & 22.86 \\
Ours & \textbf{18.91} \\
\bottomrule
\end{tabular}
\label{tab:appendix_quantitative}
\end{table}

\begin{table}[t]
    \centering
    \small
    \caption{
        User study results for perceived quality on samples generated by EP-OT-FM vs.\ OT-FM. 
        Score can range from 1 (worst) to 5 (best).
    }
    \begin{tabular}{|c|c|c|}
        \toprule
        & EP-OT-FM (Ours) & OT-FM \\
        \midrule
        Mean score      & \textbf{3.72} & 2.80 \\
        Score std. dev. & \textbf{0.24} & 0.75 \\
        \bottomrule
    \end{tabular}
    \label{tab:user_study_fm}
\end{table}

\subsection{Stroke-guided image generation (SDEdit)}
\label{sec:stroke_guided_img_gen}
We applied our edge-preserving diffusion noise to SDEdit~\cite{meng2021sdedit} for stroke-based generation. Using k-means clustering, 1000 images were converted into stroke paintings and reconstructed with backbones trained on different noise types, including blue noise~\cite{huang2024blue} and isotropic noise~\cite{ho2020denoising}, at a hijack point of $0.55T$. \textbf{Our method better adheres to guiding priors, reduces visual artifacts, and achieves superior FID scores (\cref{fig:sde_edit_comparison})}. 

Additional results (\cref{sec:appendix_additional_results}) and further evaluations on precision/recall and CLIP confirm that it maintains diversity while enhancing semantic preservation compared to the isotropic backbone. These findings highlight the usefulness of edge-preserving noise in editing tasks that rely on geometric fidelity. 

\subsection{Fine-tuning with edge-preserving noise}
We found that a model pre-trained with isotropic noise can be efficiently fine-tuned using edge-preserving noise. After fewer than 5k fine-tuning iterations on a model pre-trained for 150,000 steps (2000 epochs) on AFHQ-Cat ($128^2$), it already shows clear evidence of learning the non-isotropic noise patterns in the data (see \cref{fig:fine_tuning_noise_masks}). This improvement is reflected in the FID score, which drops from 16.03 for the pre-trained checkpoint (isotropic noise) at 2000 epochs to 12.59 after fine-tuning with edge-preserving noise. 

\section{Ablation study}
\label{sec:ablation}

\subsection{Impact of transition function $\TransitionFn$.}
We have experimented with three different choices for the transition function $\TransitionFn$: linear, cosine and sigmoid. While cosine and sigmoid show similar performance, we found that having a smooth linear transition function significantly improves the performance of the model. A qualititative and quantitative comparison between the choices is presented in \cref{fig:ablation}.

\subsection{Impact of transition points $\TransitionPt$.}
\begin{figure}[tbp]
    \centering
    \includegraphics[scale=0.5]{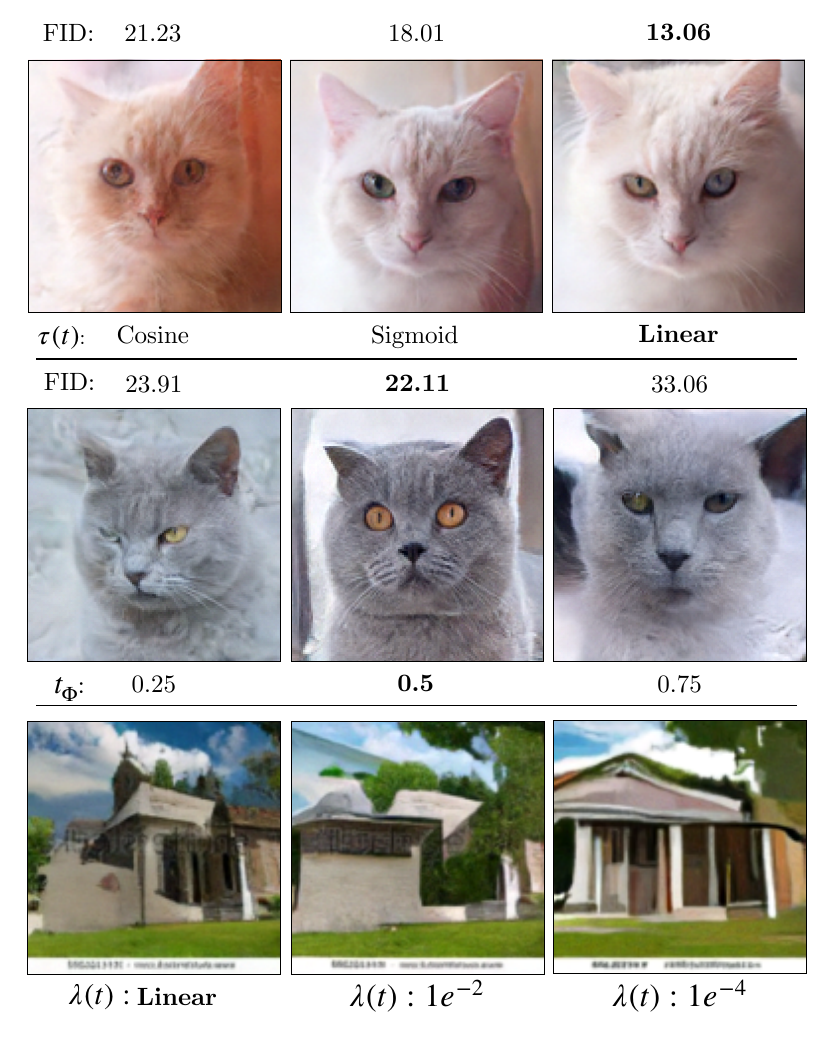}
    \caption{Impact of our noise framework's hyperparameters on sample quality. Each row shows the impact of a specific parameter (varying across 3 values) on visual sample quality and FID score. FID scores were computed on 30k samples.}
    \label{fig:ablation}
    \vspace{-3mm}
\end{figure}

{
\setlength{\columnsep}{4mm}%
\setlength{\intextsep}{3mm}%

We have investigated the impact of the transition point $\TransitionPt$ on our method's performance by considering 3 different diffusion schemes: 25\% edge-preserving - 75\% isotropic, 50\% isotropic - 50\% edge-preserving and 75\% edge-preserving - 25\% isotropic. A visual example for AFHQ-Cat ($128^2$) is presented in \cref{fig:ablation} on the right. We have experienced that there are limits to how far the transition point can be placed without sacrificing sample quality. Visually, we observe that the further the transition point is placed, the less details the model generates. The core shapes however stay intact. This is illustrated well by~\cref{fig:appendix_trans_pt_ablation} in~\cref{sec:appendix_additional_results}. For the datasets we tested on, we found that the 50\%-50\% diffusion scheme works best in terms of FID metric and visual sharpness. This again becomes apparent in ~\cref{fig:appendix_trans_pt_ablation}: although the samples for $\TransitionPt=0.25$ contain slightly more details, the samples for $\TransitionPt=0.5$ are significantly sharper. 

\subsection{Impact of edge sensitivity $\EdgeSensitivityTimeVarying$.}
 As shown in \cref{fig:ablation}, lower constant $\EdgeSensitivityTimeVarying$ values lead to less detailed, more flat, "water-painting-style" samples. Intuitively, a lower constant $\EdgeSensitivityTimeVarying$ corresponds to stronger edge-preservation in the noise and our model is explicitly trained accordingly to better learn the core structural shapes instead of the high-frequency details that we typically find in interior regions. Our time-varying choice for $\EdgeSensitivityTimeVarying$ works better than other settings in our experiments, by effectively balancing the preservation of structural information across different granularities of detail.
}

\section{Limitations and conclusion}
We introduced a new class of edge-preserving generative diffusion models that generalize isotropic models and can be applied in both the frameworks of diffusion and flow matching. Our hybrid process consists of an edge-preserving phase, which maintains structural details, followed by an isotropic phase to ensure convergence to a known prior. This decoupled approach better captures low-to-mid frequencies and accelerates convergence to sharper, less noisy predictions. While gains for unconditional image generation are present, we found the method particularly outperforms its isotropic counterpart in a shape-guided generative task. In addition, our framework offers a large hyperparameter space that remains open for further exploration. While we provided an ablation to form an intuition of each parameter's impact, we do not claim that the parameters we currently used are optimal. Future work could explore more applications that can benefit from accurate structure-guided generation, as well as experiment with our non-isotropic noise framework in video generation (e.g. for better temporal consistency).

\bibliographystyle{eg-alpha-doi} 
\bibliography{main}      

\appendix

\section{Relation to ELBO objective in diffusion literature}
\label{sec:negative_log_likelihood}
In this section we explain how the loss derivation from a perspective of minimizing the negative log-likelihood can be done for our formulation, similar to what is discussed in the original DDPM~\cite{ho2020denoising} paper.

The denoising probabilistic model paradigm defined in the DDPM paper defines the loss by minimizing a variational upper bound on the negative log likelihood. Because our noise is still Gaussian, the derivation they make in Eq. (3) to (5) of their paper still holds for us. The difference however is that we are non-isotropically scaling our noise based on the image content. As a result, our methods differ on Eq. (8) in their paper. Instead, we end up with the following form of this equation:
\begin{align}
    L_{t-1} = \mathbb{E}_q[\langle\Sigma^{-1}(\tilde{\mathbf{\mu_t}}(\mathbf{x}_t, \mathbf{x}_0) - \mathbf{\mu_{\theta}}(\mathbf{x}_t, t)) , (\tilde{\mathbf{\mu_t}}(\mathbf{x}_t, \mathbf{x}_0) - \mathbf{\mu_{\theta}}(\mathbf{x}_t, t)) \rangle]
\end{align}
In essence, for our formulation that considers non-isotropic Gaussian noise, we need to apply a different loss scaling for each pixel. A theoretically-founded \textit{weighted} version of our loss function (introduced in \cref{eq:loss_fn}) would then be the following: 

\begin{align}
    \mathcal{L} = \frac{1}{2} \langle \Sigma^{-1} (f_{\theta}(\mathbf{x}_t, t)-\mathbf{\sigma}_t \mathbf{\epsilon}_t)) , (f_{\theta}(\mathbf{x}_t, t)-\mathbf{\sigma}_t \mathbf{\epsilon}_t) \rangle
\end{align}

However, note that in the original DDPM paper, the loss is also simplified by removing the weighting, which they call the simplified or \textit{reweighted loss} (Eq. (14) in their paper). The authors argue that this reweighting leads to improved sample quality. To save computational resources, we follow a similar heuristic in our loss function where we remove the weighting. While our heuristic loss function already proved effective, a more theoretically accurate loss would include the scaling discussed above. 

In our non-isotropic case, $\Sigma$ is dependent on both the clean data $\Samplezero$ and $\timepoint$. Therefore, the scaling could be approximated by choosing $\hat{\Sigma_{\timepoint}}$ such that $c_1||\hat{\Sigma_{\timepoint}}|| \le ||\Sigma(\Samplezero)_{\timepoint}|| \le c_2||\hat{\Sigma_{\timepoint}}||$, for some $c_1, c_2 > 0$ and for all $\Samplezero$, where $|| . ||$ is an appropriate norm.

\begin{table}[t]
\small
\centering
\caption{
Average FID score over first 10k training iterations for our edge-preserving model and the isotropic model DDPM trained on different frequency bands of the AFHQ-Cat($128^2$) dataset. $\sigma$ corresponds to the standard deviation of the Gaussian filtering kernel used to obtain a certain frequency band of the dataset.
}

\begin{tabular}{|c|c|c|c|c|c|c|}
\toprule
FID ($\downarrow$) & $\sigma=1.0$ & $\sigma=2.5$ & $\sigma=5.0$ & $\sigma=7.5$ & $\sigma=10.0$ \\
\midrule
DDPM & \textbf{58.64} & 100.78 & 204.40 & 265.55 & 291.76 \\
Ours & 76.65 & \textbf{78.97} & \textbf{138.55} & \textbf{194.35} & \textbf{250.88}  \\
\bottomrule
\end{tabular}
\label{tab:freq_analysis_results_table}
\end{table}

\section{Analysis of model's capacity across different frequency bands}
\label{sec:frequency_analysis}
To better understand the impact of edge-preserving noise on modeling the target distribution, we conducted an analysis on its training performance for different frequency bands. Our setup is as follows, we create 5 versions of the AFHQ-Cat128 dataset, each with a different cutoff frequency. This corresponds to convoluting each image in the dataset with a Gaussian kernel of a specific standard deviation $\sigma$, representing a frequency band (as $\sigma$ increases, fewer high-frequency components are preserved). For each frequency band, we then trained a model with edge-preserving noise and a model with isotropic noise (DDPM) for a fixed amount of 10000 training iterations. We place a model checkpoint at every 1000 iterations, and computed the average FID score ($N=1000$ samples) per frequency band over all these checkpoints. We present the corresponding results in \cref{tab:freq_analysis_results_table}. These preliminary results suggest that edge-preserving noise improves learning of low-to-mid frequencies compared to isotropic noise.


\section{Implementation details of experiments}
\label{sec:exp_implementation_details}
We compare our method against two baselines that use an isotropic form of noise, namely DDPM~\cite{ho2020denoising} and Optimal Transport Flow Matching (OT-FM) \cite{lipman2022flow}.

We perform unconditional generation experiments on two settings: pixel-space diffusion following the setting of ~\cite{ho2020denoising,rissanen2022generative} and latent-space diffusion following~\cite{rombach2022high} noted as LDM in~\cref{tab:appendix_quantitative}, where the diffusion process runs in the latent space. We also experimented with edge-preserving noise applied to the framework of flow matching. Besides this, we perform an experiment on shape-guided generation, as summarized in \cref{fig:sde_edit_comparison}, and an analysis on the capabilities of our model to learn different frequency bands of the data, further explained in \cref{sec:frequency_analysis}.
We used the following datasets: CIFAR-10 ($32^2$, 50,000 training images)~\cite{krizhevsky2009learning}, CelebA ($64^2$ (flow matching), $128^2$ (diffusion), 30,000 training images)~\cite{CelebAMask-HQ}, AFHQ-Cat ($128^2$, 5,153 training images)~\cite{choi2020stargan}, Human-Sketch ($128^2$, 20,000 training images) ~\cite{eitz2012hdhso} (see ~\cref{fig:appendix_human_sketch_dataset}) and LSUN-Church ($128^2$, 126,227 training images)~\cite{yu2015lsun} for pixel-space diffusion and/or flow matching. For latent-space diffusion~\cite{rombach2022high}, we tested on AFHQ-Cat ($512^2$). 

We used a batch size of 64 for all experiments in image space, and a batch size of 128 for all experiments in latent space.
We trained CIFAR-10($32^2$) and AFHQ-Cat ($128^2$) for 1000 epochs, AFHQ-Cat ($512^2$) (latent diffusion) for 1750 epochs, CelebA($128^2$) for 475 epochs and LSUN-Church($128^2$) for 90 epochs for our method and the baselines we compare to.
Our framework is implemented in Pytorch~\cite{paszke2017automatic}. 
For the network architecture we adopt the 2D U-Net from~\cite{rissanen2022generative}.
We use T = 500 discrete time steps for both training and inference, except for AFHQ-Cat ($128^2$), where we used T = 750.
To optimize the network parameters, we use Adam optimizer~\cite{kingma2014adam} with learning rate $1e^{-4}$ for latent-space diffusion models and $2e^{-5}$ for pixel-space diffusion models. We trained all datasets on 2x NVIDIA Tesla A40.

For our final results, we used a linear scheme for $\EdgeSensitivityTimeVarying$ that linearly interpolates between $\EdgeSensitivityMin = 1e^{-4}$ and $\EdgeSensitivityMax = 1e^{-1}$. We used a transition point $\TransitionPt = 0.5$ and a linear transition function $\TransitionFn$.

To evaluate the quality of generated samples, we consider FID~\cite{heusel2017gans}.
using the implementation from~\cite{stein2024exposing}, with Inception-v3 network~\cite{szegedy2016rethinking} as backbone.
We generated 30k images to compute FID scores for unconditional generation and shape-guided generation, for all datasets.

\begin{figure}[tbp]
    \centering
    \includegraphics[width=0.7\linewidth]{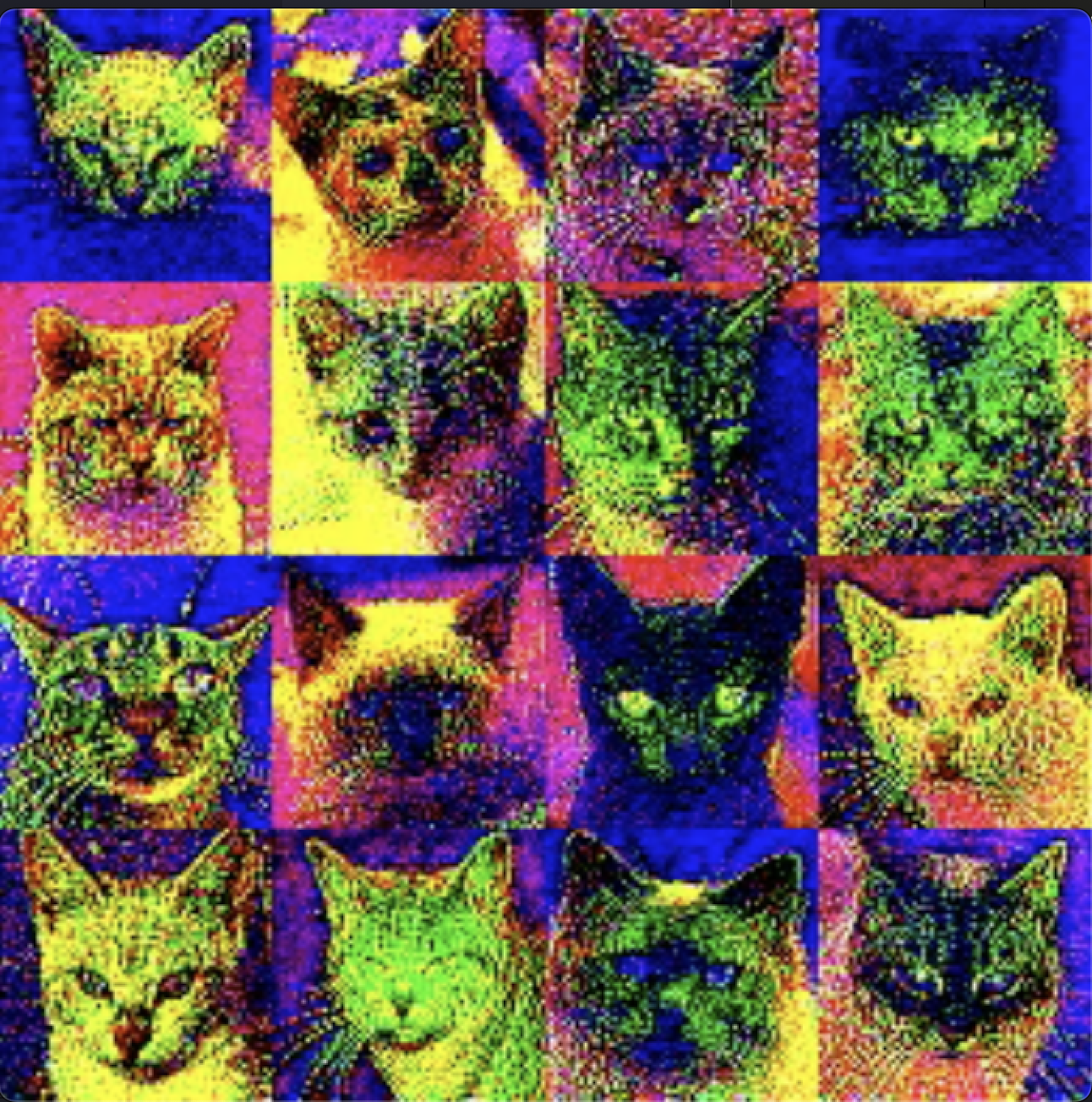}
    \caption{We visualize image latents for the AFHQ-Cat($512^2$) dataset. Notice that most of the structural content of the image remains preserved in the latent space. Therefore, it makes sense to also apply edge-preserving noise in the latent space (also see \cref{tab:appendix_quantitative} and \cref{fig:appendix_cat_res512}).}
    \label{fig:latents_visualization}
\end{figure}

\section{Additional results}
\label{sec:appendix_additional_results}

In this section, we provide additional results and ablations.

\Cref{tab:appendix_quantitative} shows quantitative FID comparisons using latent diffusion \cite{rombach2022high} models on all the baselines. \Cref{fig:appendix_cat_res128}, \Cref{fig:appendix_celeba_res128},  \Cref{fig:appendix_church_res128},  \Cref{fig:appendix_cat_res512} show more generated samples and comparisons with DDPM on all previously introduced datasets.
In ~\cref{fig:appendix_human_sketch_dataset} we show  samples for the Human-Sketch $(128^2)$ data set specifically. This dataset was of particular interest to us, given the images only consist of high-frequency, edge content. Although we observed that this data is
remarkably challenging for all methods, our model is able to consistently deliver visually better results. 

 \Cref{fig:appendix_trans_pt_ablation} shows an additional visualization of the impact $\TransitionPt$ for the LSUN-Church ($128^2$) dataset. $\TransitionPt=0.5$ works best in terms of FID metric, consistent to the results shown in \cref{sec:ablation}.

\begin{figure}[H]
    \centering

\newcommand{\PlotSingleImage}[1]{%
        \begin{scope}
            \clip (0,0) -- (2.5,0) -- (2.5,2.5) -- (0,2.5) -- cycle;
            \path[fill overzoom image=figures/#1] (0,0) rectangle (2.5cm,2.5cm);
        \end{scope}
        \draw (0,0) -- (2.5,0) -- (2.5,2.5) -- (0,2.5) -- cycle;
        
}

\newcommand{\PlotSingleImageWithLine}[1]{%
        \begin{scope}
            \clip (0,0) -- (2.5,0) -- (2.5,2.5) -- (0,2.5) -- cycle;
            \path[fill overzoom image=figures/#1] (0,0) rectangle (2.5cm,2.5cm);
        \end{scope}
        \draw (0,0) -- (2.5,0) -- (2.5,2.5) -- (0,2.5) -- cycle;
        \draw[dashed] (0, -0.13) -- (2.5, -0.13);
}

\newcommand{\TwoColumnFigure}[2]{%
    \begin{tabular}{c@{\;}c@{}}
        \hspace*{-2.5mm}
        \begin{tikzpicture}[scale=0.563]
            \PlotSingleImage{#1}
        \end{tikzpicture}
         & 
         \begin{tikzpicture}[scale=0.563]
            \PlotSingleImage{#2}
        \end{tikzpicture}
    \end{tabular}%
}
\newcommand\scalevalue{0.64}    

%
\hspace*{-5mm}
\begin{tabular}{c@{\;}c@{}}
\footnotesize
\begin{tabular}{c@{\;}c@{\;}c@{\;}c@{\;}c@{\;}c@{}}

\rotatebox{90}{\tiny DDPM (FID: 67.96)}
&
\begin{tikzpicture}[scale=\scalevalue]
\PlotSingleImage{human_sketch_dataset/ddpm/1.png}
\end{tikzpicture}
&
\begin{tikzpicture}[scale=\scalevalue]
\PlotSingleImage{human_sketch_dataset/ddpm/2.png}
\end{tikzpicture}
&
\begin{tikzpicture}[scale=\scalevalue]
\PlotSingleImage{human_sketch_dataset/ddpm/3.png}
\end{tikzpicture}
&
\begin{tikzpicture}[scale=\scalevalue]
\PlotSingleImage{human_sketch_dataset/ddpm/5.png}
\end{tikzpicture}
&
\begin{tikzpicture}[scale=\scalevalue]
\PlotSingleImage{human_sketch_dataset/ddpm/6.png}
\end{tikzpicture}
\\
\rotatebox{90}{ \tiny \textbf{Ours (FID: 40.03)}}
&
\begin{tikzpicture}[scale=\scalevalue]
\PlotSingleImage{human_sketch_dataset/ours/6.png}
\end{tikzpicture}
&
\begin{tikzpicture}[scale=\scalevalue]
\PlotSingleImage{human_sketch_dataset/ours/2.png}
\end{tikzpicture}
&
\begin{tikzpicture}[scale=\scalevalue]
\PlotSingleImage{human_sketch_dataset/ours/3.png}
\end{tikzpicture}
&
\begin{tikzpicture}[scale=\scalevalue]
\PlotSingleImage{human_sketch_dataset/ours/4.png}
\end{tikzpicture}
&
\begin{tikzpicture}[scale=\scalevalue]
\PlotSingleImage{human_sketch_dataset/ours/5.png}
\end{tikzpicture}
\\
\end{tabular}
&
\end{tabular} 
    \caption{Generated unconditional samples for the Human Sketch ($128^2$) dataset \cite{eitz2012hdhso}. Both models were trained for an equal amount of 575 epochs.}
     \label{fig:appendix_human_sketch_dataset}
\end{figure}

\begin{figure*}[tbp]
    \centering

\newcommand{\PlotSingleImage}[1]{%
        \begin{scope}
            \clip (0,0) -- (2.5,0) -- (2.5,2.5) -- (0,2.5) -- cycle;
            \path[fill overzoom image=figures/#1] (0,0) rectangle (2.5cm,2.5cm);
        \end{scope}
        \draw (0,0) -- (2.5,0) -- (2.5,2.5) -- (0,2.5) -- cycle;
        
}

\newcommand{\PlotSingleImageWithLine}[1]{%
        \begin{scope}
            \clip (0,0) -- (2.5,0) -- (2.5,2.5) -- (0,2.5) -- cycle;
            \path[fill overzoom image=figures/#1] (0,0) rectangle (2.5cm,2.5cm);
        \end{scope}
        \draw (0,0) -- (2.5,0) -- (2.5,2.5) -- (0,2.5) -- cycle;
        \draw[dashed] (0, -0.13) -- (2.5, -0.13);
}

\newcommand{\TwoColumnFigure}[2]{%
    \begin{tabular}{c@{\;}c@{}}
        \hspace*{-2.5mm}
        \begin{tikzpicture}[scale=0.563]
            \PlotSingleImage{#1}
        \end{tikzpicture}
         & 
         \begin{tikzpicture}[scale=0.563]
            \PlotSingleImage{#2}
        \end{tikzpicture}
    \end{tabular}%
}
\newcommand\scalevalue{0.61}    
\newcommand\smallerscale{0.55}    

\hspace*{-5mm}
\begin{tabular}{c@{\;}c@{}}
\footnotesize
\begin{tabular}{c@{\;}c@{\;}c@{\;}c@{\;}c@{}}


&
\rotatebox{90}{ \tiny CelebA ($128^2$)}
\begin{tikzpicture}[scale=\scalevalue]
\PlotSingleImage{sde_edit_comparison/celeba/extras/painting_458.png}
\end{tikzpicture}
&
\begin{tikzpicture}[scale=\scalevalue]
\PlotSingleImage{sde_edit_comparison/celeba/extras/bndm_458.png}
\end{tikzpicture}

&
\begin{tikzpicture}[scale=\scalevalue]
\PlotSingleImage{sde_edit_comparison/celeba/extras/ddpm_458.png}
\end{tikzpicture}
&
\begin{tikzpicture}[scale=\scalevalue]
\PlotSingleImage{sde_edit_comparison/celeba/extras/ours_458.png}
\end{tikzpicture}
\\
&
\rotatebox{90}{\hspace{1mm}\tiny Church ($128^2$)}
\begin{tikzpicture}[scale=\scalevalue]
\PlotSingleImage{sde_edit_comparison/church/extras/painting_476.png}
\end{tikzpicture}
&
\begin{tikzpicture}[scale=\scalevalue]
\PlotSingleImage{sde_edit_comparison/church/extras/bndm_476.png}
\end{tikzpicture}
&
\begin{tikzpicture}[scale=\scalevalue]
\PlotSingleImage{sde_edit_comparison/church/extras/ddpm_476.png}
\end{tikzpicture}
&
\begin{tikzpicture}[scale=\scalevalue]
\PlotSingleImage{sde_edit_comparison/church/extras/ours_476.png}
\end{tikzpicture}
\\
&
\rotatebox{90}{\hspace{3mm}\tiny Cat ($128^2$)}
\begin{tikzpicture}[scale=\scalevalue]
\PlotSingleImage{sde_edit_comparison/cat/extras/painting_229.png}
\end{tikzpicture}
&
\begin{tikzpicture}[scale=\scalevalue]
\PlotSingleImage{sde_edit_comparison/cat/extras/bndm_229.png}
\end{tikzpicture}
&
\begin{tikzpicture}[scale=\scalevalue]
\PlotSingleImage{sde_edit_comparison/cat/extras/ddpm_229.png}
\end{tikzpicture}
&
\begin{tikzpicture}[scale=\scalevalue]
\PlotSingleImage{sde_edit_comparison/cat/extras/ours_229.png}
\end{tikzpicture}
\\
& Synthetic & Blue noise & Isotropic & \textbf{Ours} \\ & painting &  &  & 
\\
\end{tabular}
&
\hspace{0.025cm}

\begin{tabular}{c@{\;}c@{\;}c@{\;}c@{\;}c@{}}


&
\begin{tikzpicture}[scale=\scalevalue]
\PlotSingleImage{sde_edit_comparison/celeba/extras/painting_78.png}
\end{tikzpicture}
&
\begin{tikzpicture}[scale=\scalevalue]
\PlotSingleImage{sde_edit_comparison/celeba/extras/bndm_78.png}
\end{tikzpicture}

&
\begin{tikzpicture}[scale=\scalevalue]
\PlotSingleImage{sde_edit_comparison/celeba/extras/ddpm_78.png}
\end{tikzpicture}
&
\begin{tikzpicture}[scale=\scalevalue]
\PlotSingleImage{sde_edit_comparison/celeba/extras/ours_78.png}
\end{tikzpicture}
\\
&
\begin{tikzpicture}[scale=\scalevalue]
\PlotSingleImage{sde_edit_comparison/church/extras/painting_317.png}
\end{tikzpicture}
&
\begin{tikzpicture}[scale=\scalevalue]
\PlotSingleImage{sde_edit_comparison/church/extras/bndm_317.png}
\end{tikzpicture}
&
\begin{tikzpicture}[scale=\scalevalue]
\PlotSingleImage{sde_edit_comparison/church/extras/ddpm_317.png}
\end{tikzpicture}
&
\begin{tikzpicture}[scale=\scalevalue]
\PlotSingleImage{sde_edit_comparison/church/extras/ours_317.png}
\end{tikzpicture}
\\
&
\begin{tikzpicture}[scale=\scalevalue]
\PlotSingleImage{sde_edit_comparison/cat/extras/painting_326.png}
\end{tikzpicture}
&
\begin{tikzpicture}[scale=\scalevalue]
\PlotSingleImage{sde_edit_comparison/cat/extras/bndm_326.png}
\end{tikzpicture}
&
\begin{tikzpicture}[scale=\scalevalue]
\PlotSingleImage{sde_edit_comparison/cat/extras/ddpm_326.png}
\end{tikzpicture}
&
\begin{tikzpicture}[scale=\scalevalue]
\PlotSingleImage{sde_edit_comparison/cat/extras/ours_326.png}
\end{tikzpicture}
\\
& Synthetic & Blue noise & Isotropic & \textbf{Ours} \\ & painting &  &  & 
\\
\end{tabular}


\end{tabular} 
    \caption{More samples for our model and other baselines applied to SDEdit \cite{meng2021sdedit}. Note how our model is able to generate sharper results that suffer less from artifacts. Although BNDM can generate satisfactory results in certain cases (e.g., cat and church), it often deviates from the stroke painting guide, potentially producing outcomes that differ significantly from the user's original intent. In contrast, our method closely follows the stroke painting guide, accurately preserving both shape and color.}
     \label{fig:appendix_sdedit}
\end{figure*}

\begin{figure*}[tbp]
    \centering

\newcommand{\PlotSingleImage}[1]{%
        \begin{scope}
            \clip (0,0) -- (2.5,0) -- (2.5,2.5) -- (0,2.5) -- cycle;
            \path[fill overzoom image=figures/#1] (0,0) rectangle (2.5cm,2.5cm);
        \end{scope}
        \draw (0,0) -- (2.5,0) -- (2.5,2.5) -- (0,2.5) -- cycle;
        
}

\newcommand{\PlotSingleImageNonSquared}[1]{%
        \begin{scope}
            \clip (0,0) -- (4,0) -- (4,2) -- (0,2) -- cycle;
            \path[fill overzoom image=figures/#1] (0,0) rectangle (4cm,2cm);
        \end{scope}
        \draw (0,0) -- (4,0) -- (4,2) -- (0,2) -- cycle;
        
}



\newcommand\scalevalue{1.07}    

%
\hspace{-0.8cm}
\begin{tabular}{c@{\;}c@{\;}c@{}}
FID: $39.31$ & $\textbf{35.71}$ & $86.41$\\
\begin{tabular}{c@{\;}c@{}}
\begin{tikzpicture}[scale=\scalevalue]
\PlotSingleImageNonSquared{appendix_trans_pt_ablation/0pt25.png}
\end{tikzpicture}
\\[-0.4mm]
0.25
\\[-0.4mm]
\end{tabular}
&
\begin{tabular}{c@{\;}c@{}}
\begin{tikzpicture}[scale=\scalevalue]
\PlotSingleImageNonSquared{appendix_trans_pt_ablation/0pt5.png}
\end{tikzpicture}
\\[-0.4mm]
0.5
\\[-0.4mm]
\end{tabular}&
\begin{tabular}{c@{\;}c@{}}
\begin{tikzpicture}[scale=\scalevalue]
\PlotSingleImageNonSquared{appendix_trans_pt_ablation/0pt75.png}
\end{tikzpicture}
\\[-0.4mm]
0.75
\\[-0.4mm]
\end{tabular}
\end{tabular}
 
    \caption{
        Impact of location of transition point $\TransitionPt$ on sample quality, shown for the LSUN-Church ($128^2$) dataset. If we place $\TransitionPt$ too far, the model happens to learn only the lowest frequencies and generates no details at all. Placing it too early leads to results that are less sharp. We found that by placing $\TransitionPt$ at 50\%, we strike a good balance between the two, leading to better quantitative and qualitative results.
    }
    \label{fig:appendix_trans_pt_ablation}
\end{figure*}

\vspace{3mm}
\begin{table*}[ht]
\small
\centering
\caption{
Shape-guided image generation (based on SDEdit~\cite{meng2021sdedit}): precision (metric for realism) and recall (metric for diversity) scores~\cite{kynkaanniemi2019improved} for isotropic model DDPM, and our edge-preserving model.
We consistently outperform in terms of precision, and closely match in terms of recall.
}

\begin{tabular}{|c|cc|cc|}
\toprule
& \multicolumn{2}{c|}{Ours} & \multicolumn{2}{c|}{Isotropic noise} \\
Shape-guided image generation & Precision ($\uparrow$) & Recall ($\uparrow$) & Precision ($\uparrow$) & Recall ($\uparrow$) \\
\hline
AFHQ-Cat($128^2$) & \textbf{0.93} & \textbf{0.80} & 0.92 & 0.66 \\
CelebA($128^2$) & \textbf{0.65} & 0.46 & 0.53 & \textbf{0.53} \\
LSUN-Church($128^2$) & \textbf{0.87} & 0.46 & 0.84 & \textbf{0.50} \\
\bottomrule
\end{tabular}
\label{tab:shape_guided_precision_and_recall}
\end{table*}

\vspace{3mm}
\begin{table*}[ht]
\small
\centering
\caption{
Unconditional image generation: precision (metric for realism) and recall (metric for diversity) scores for isotropic model DDPM, and our edge-preserving model.
While our model slightly get outperformed, we find that our edge-preserving model closely matches DDPM on both metrics. We would therefore argue that edge-preserving noise minimally impacts diversity.
}

\begin{tabular}{|c|cc|cc|}
\toprule
& \multicolumn{2}{c|}{Ours} & \multicolumn{2}{c|}{Isotropic noise} \\
Unconditional image generation & Precision ($\uparrow$) & Recall ($\uparrow$) & Precision ($\uparrow$) & Recall ($\uparrow$) \\
\hline
AFHQ-Cat($128^2$) & 0.76 & 0.20 & \textbf{0.77} & \textbf{0.21} \\
CelebA($128^2$) & 0.90 & 0.16 & \textbf{0.92} & \textbf{0.17} \\
LSUN-Church($128^2$) & \textbf{0.65} & 0.33 & 0.47 & \textbf{0.38} \\
\bottomrule
\end{tabular}
\label{tab:unconditional_precision_and_recall}
\end{table*}

\vspace{3mm}

\begin{table}[H]
    \small
    \centering
\caption{
Additional CLIP-score comparisons~\cite{radford2021learning} for stroke-guided generation~\cite{meng2021sdedit} show that our method consistently outperforms the isotropic baseline, producing images that are more semantically aligned with the originals.
}

\begin{tabular}{|c|c|c|c|}
\toprule
CLIP-score & Ours & DDPM \\
\midrule
AFHQ-Cat($128^2$) & \textbf{88.97} & 88.78 \\
CelebA($128^2$) & \textbf{61.15} & 61.02  \\
LSUN-Church($128^2$) & \textbf{64.32} & 62.57 \\
\bottomrule
\end{tabular}
\label{tab:shape_guided_clip}
\end{table}
\vspace{3mm}

\begin{figure}[H]
    \centering

\newcommand{\PlotSingleImage}[1]{%
        \begin{scope}
            \clip (0,0) -- (2.5,0) -- (2.5,2.5) -- (0,2.5) -- cycle;
            \path[fill overzoom image=figures/#1] (0,0) rectangle (2.5cm,2.5cm);
        \end{scope}
        \draw (0,0) -- (2.5,0) -- (2.5,2.5) -- (0,2.5) -- cycle;
        
}

\newcommand{\PlotSingleImageNonSquared}[1]{%
        \begin{scope}
            \clip (0,0) -- (4,0) -- (4,10) -- (0,10) -- cycle;
            \path[fill overzoom image=figures/#1] (0,0) rectangle (4cm,10cm);
        \end{scope}
        \draw (0,0) -- (4,0) -- (4,10) -- (0,10) -- cycle;
        
}



\newcommand\scalevalue{0.90}    

%
\hspace*{-4.0mm}
\begin{tabular}{c@{\;}c@{\;}c@{\;}c@{}}
&
\begin{tabular}{c@{\;}c@{}}
\begin{tikzpicture}[scale=\scalevalue]
\PlotSingleImageNonSquared{celeba_res128/ddpm.png}
\end{tikzpicture}
\\[-0.4mm]
DDPM
\\[-0.4mm]
\end{tabular}
\begin{tabular}{c@{\;}c@{}}
\begin{tikzpicture}[scale=\scalevalue]
\PlotSingleImageNonSquared{celeba_res128/ours.png}
\end{tikzpicture}
\\[-0.4mm]
Ours
\\[-0.4mm]
\end{tabular}
\end{tabular}
 
    \caption{
        More unconditional samples for DDPM (isotropic noise) and our edge-preserving noise on the CelebA ($128^2$) dataset.
    }
    \label{fig:appendix_celeba_res128}
\end{figure}
\begin{figure}[H]
    \centering

\newcommand{\PlotSingleImage}[1]{%
        \begin{scope}
            \clip (0,0) -- (2.5,0) -- (2.5,2.5) -- (0,2.5) -- cycle;
            \path[fill overzoom image=figures/#1] (0,0) rectangle (2.5cm,2.5cm);
        \end{scope}
        \draw (0,0) -- (2.5,0) -- (2.5,2.5) -- (0,2.5) -- cycle;
        
}

\newcommand{\PlotSingleImageNonSquared}[1]{%
        \begin{scope}
            \clip (0,0) -- (4,0) -- (4,10) -- (0,10) -- cycle;
            \path[fill overzoom image=figures/#1] (0,0) rectangle (4cm,10cm);
        \end{scope}
        \draw (0,0) -- (4,0) -- (4,10) -- (0,10) -- cycle;
        
}



\newcommand\scalevalue{0.90}    

%
\hspace*{-4.0mm}
\begin{tabular}{c@{\;}c@{\;}c@{\;}c@{}}
&
\begin{tabular}{c@{\;}c@{}}
\begin{tikzpicture}[scale=\scalevalue]
\PlotSingleImageNonSquared{church_res_128/ddpm.png}
\end{tikzpicture}
\\[-0.4mm]
DDPM
\\[-0.4mm]
\end{tabular}
\begin{tabular}{c@{\;}c@{}}
\begin{tikzpicture}[scale=\scalevalue]
\PlotSingleImageNonSquared{church_res_128/ours.png}
\end{tikzpicture}
\\[-0.4mm]
Ours
\\[-0.4mm]
\end{tabular}
\end{tabular}
 
    \caption{
        More unconditional samples for DDPM (isotropic noise) and our edge-preserving noise on the LSUN-Church ($128^2$) dataset. Although our results appear similar to DDPM's, our method more effectively captures the geometric details of buildings and exhibits fewer artifacts, such as blurry regions, compared to DDPM.
    }
    \label{fig:appendix_church_res128}
\end{figure}

\begin{figure}[H]
    \centering

\newcommand{\PlotSingleImage}[1]{%
        \begin{scope}
            \clip (0,0) -- (2.5,0) -- (2.5,2.5) -- (0,2.5) -- cycle;
            \path[fill overzoom image=figures/#1] (0,0) rectangle (2.5cm,2.5cm);
        \end{scope}
        \draw (0,0) -- (2.5,0) -- (2.5,2.5) -- (0,2.5) -- cycle;
        
}

\newcommand{\PlotSingleImageNonSquared}[1]{%
        \begin{scope}
            \clip (0,0) -- (4,0) -- (4,10) -- (0,10) -- cycle;
            \path[fill overzoom image=figures/#1] (0,0) rectangle (4cm,10cm);
        \end{scope}
        \draw (0,0) -- (4,0) -- (4,10) -- (0,10) -- cycle;
        
}



\newcommand\scalevalue{0.90}    

%
\hspace*{-4.0mm}
\begin{tabular}{c@{\;}c@{\;}c@{\;}c@{}}
&
\begin{tabular}{c@{\;}c@{}}
\begin{tikzpicture}[scale=\scalevalue]
\PlotSingleImageNonSquared{cat_res128/ddpm.png}
\end{tikzpicture}
\\[-0.4mm]
DDPM
\\[-0.4mm]
\end{tabular}
&
\begin{tabular}{c@{\;}c@{}}
\begin{tikzpicture}[scale=\scalevalue]
\PlotSingleImageNonSquared{cat_res128/ours.png}
\end{tikzpicture}
\\[-0.4mm]
Ours
\\[-0.4mm]
\end{tabular}
\end{tabular}
 
    \caption{
        More unconditional samples for DDPM (isotropic noise) and our edge-preserving noise on the AFHQ-Cat ($128^2$) dataset.
    }
    \label{fig:appendix_cat_res128}
\end{figure}
\begin{figure}[H]
    \centering

\newcommand{\PlotSingleImage}[1]{%
        \begin{scope}
            \clip (0,0) -- (2.5,0) -- (2.5,2.5) -- (0,2.5) -- cycle;
            \path[fill overzoom image=figures/#1] (0,0) rectangle (2.5cm,2.5cm);
        \end{scope}
        \draw (0,0) -- (2.5,0) -- (2.5,2.5) -- (0,2.5) -- cycle;
        
}

\newcommand{\PlotSingleImageNonSquared}[1]{%
        \begin{scope}
            \clip (0,0) -- (4,0) -- (4,10) -- (0,10) -- cycle;
            \path[fill overzoom image=figures/#1] (0,0) rectangle (4cm,10cm);
        \end{scope}
        \draw (0,0) -- (4,0) -- (4,10) -- (0,10) -- cycle;
        
}



\newcommand\scalevalue{0.90}    

%
\hspace*{-4.0mm}
\begin{tabular}{c@{\;}c@{\;}c@{}}
\begin{tabular}{c@{\;}c@{}}
\begin{tikzpicture}[scale=\scalevalue]
\PlotSingleImageNonSquared{ldm_cat_res512/ddpm_cat_512_grid_appendix_3.png}
\end{tikzpicture}
\\[-0.4mm]
DDPM
\\[-0.4mm]
\end{tabular}
\begin{tabular}{c@{\;}c@{}}
\begin{tikzpicture}[scale=\scalevalue]
\PlotSingleImageNonSquared{ldm_cat_res512/ours_cat_512_grid_appendix_3.png}
\end{tikzpicture}
\\[-0.4mm]
Ours
\\[-0.4mm]
\end{tabular}
\end{tabular}
 
    \caption{
        More unconditional samples for DDPM (isotropic noise) and our edge-preserving noise on the AFHQ-Cat ($512^2$, LDM) dataset. All samples are generated via diffusion in latent space. While difference in visual quality is subtle, \cref{tab:appendix_quantitative} shows that our noise framework improved the FID score.
    }
    \label{fig:appendix_cat_res512}
\end{figure}

\begin{figure}[tbp]
    \centering
    \includegraphics[width=1.0\linewidth]{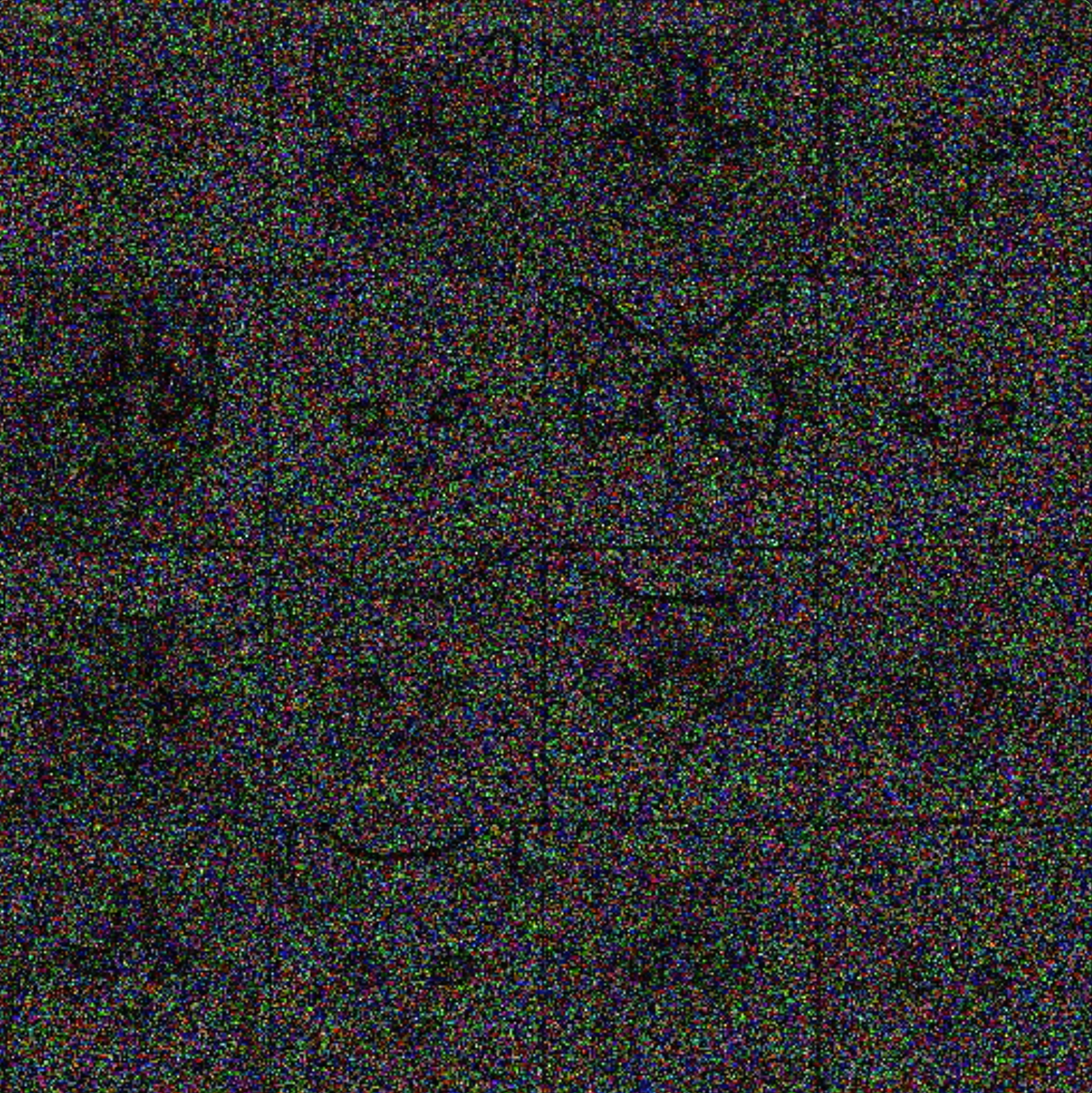}
    \vspace{0.3mm}
    \caption{Grid of predicted noises for a batch of 16 samples after fine-tuning a model pre-trained with isotropic noise for 2000 epochs on the AFHQ-Cat ($128^2$) dataset. After fewer than 5k fine-tuning iterations with edge-preserving noise, the model has already learned the non-isotropic variance corresponding to the structures in the data.}
    \label{fig:fine_tuning_noise_masks}
\end{figure}


\end{document}